\documentclass[11pt]{article}

\usepackage[preprint]{latex/acl}

\usepackage[T1]{fontenc}
\usepackage[dvipsnames,svgnames]{xcolor}
\usepackage{tikz}
\usepackage{tikz-3dplot}
\usetikzlibrary{3d,positioning, shapes.geometric, decorations.pathmorphing, backgrounds, fit, calc, decorations.pathreplacing,arrows.meta,calligraphy}
\usetikzlibrary{shadows, automata, positioning}
\usepackage{times}
\usepackage{latexsym}
\usepackage[utf8]{inputenc}
\usepackage{microtype}
\usepackage{inconsolata}
\usepackage{amssymb}
\usepackage{amsmath}
\usepackage{multirow}
\usepackage{adjustbox}
\usepackage{svg}
\usepackage{pgfplots}
\pgfplotsset{compat=1.18}

\usepackage[disable]{todonotes}
\usepackage{booktabs}
\usepackage{xspace}
\usepackage{graphicx}
\usepackage{float}
\usepackage{placeins}

\usepackage{bm}
\usepackage{todonotes}
\todostyle{jlr}{color=red,shadow}

\usepackage{comment}

\usepackage{mathtools}
\DeclareMathOperator*{\E}{\mathbb{E}}
\newcommand{\KL}{D_{\mathrm{KL}}}
\newcommand{\sys}[1]{\textsc{#1}}

\newcommand{\KLsplit}[2]{%
  \KL\!\vcenter{\hbox{$
    \left(
      \begin{array}{@{}l@{}}
        #1 \\
        [.6ex] || \; #2
      \end{array}
    \right)$}}%
}

\usepackage[linesnumbered,ruled,vlined]{algorithm2e}
\usepackage{algpseudocode}

\makeatletter
\DeclareRobustCommand\onedot{\futurelet\@let@token\@onedot}
\def\@onedot{\ifx\@let@token.\else.\null\fi\xspace}

\def\ie{\emph{i.e}\onedot} 
\def\cf{\emph{cf}\onedot}

\makeatother

\usepackage{amsmath,amsfonts,bm}

\def\eqref#1{equation~\ref{#1}}

\def\1{\bm{1}}

\DeclareMathAlphabet{\mathsfit}{\encodingdefault}{\sfdefault}{m}{sl}
\SetMathAlphabet{\mathsfit}{bold}{\encodingdefault}{\sfdefault}{bx}{n}

\DeclareMathOperator*{\argmax}{argmax}

\title{Diffusion-MF: Approximate Structured Diffusion for Sequence Labelling}

\author{Nicolas Floquet, Joseph Le~Roux, Nadi Tomeh\\
 Université Sorbonne Paris Nord, CNRS, \\ Laboratoire d'Informatique de Paris Nord, \\LIPN, F-93430 Villetaneuse, France \\
  \texttt{\{floquet, leroux, tomeh\}@lipn.fr} \\}

\begin{document}
\maketitle
\begin{abstract}
We introduce \sys{Diffusion-MF}, a discrete diffusion sequence labeller
that places a linear-chain conditional random field (LCRF) inside the denoising loop.
Unlike prior diffusion labellers
, it performs
structured inference at every step; parallel Mean-Field makes this efficient.
Across multilingual POS, CoNLL-2003 NER, and Chinese Seg+POS,
\sys{Diffusion-MF} achieves the best primary result across all six experimental
settings.\footnote{Code is available at
\url{https://anonymous.4open.science/r/diffusion-mf-artifact-1058/}.}
\end{abstract}

\section{Introduction}
\label{sec:introduction}

Sequence labelling maps a sentence to labels, as in POS tagging and
named-entity recognition.  Linear-chain conditional random fields (LCRFs)
explicitly model adjacent labels and support efficient inference
\citep{lafferty-2001-condit-random-field,zheng-2015-condit-random,
zhang-etal-2020-efficient}.
Although neural encoders provide global input context, an LCRF models output
dependencies only between adjacent labels; arbitrary long-range factors would
forfeit efficient inference.  We instead seek global interactions among
predicted labels while retaining explicit transitions and efficient inference.

Diffusion provides a principled mechanism for global interaction by
iteratively refining a complete noisy sequence: a Transformer denoiser can
condition every update on the entire evolving prediction.  In sequence
labelling, structure is defined over categorical tags and their transitions;
keeping the diffusion process in label space lets the denoiser model
this structure directly.  Discrete diffusion provides such a process
\citep{hoogeboom-2021-argmax-flows,austin-2021-struc-denois}, yet existing
diffusion sequence labellers use continuous relaxations.
\sys{DiffusionSL} maps tags to bit vectors, applies Gaussian noise, and
thresholds the resulting vectors back into labels
\citep{huang-etal-2023-diffusionsl}.  This imposes an artificial geometry
over tags and lacks categorical unary or sequence
scores, making explicit constraints in structured decoding difficult.

To our knowledge, we introduce the first diffusion sequence labeller whose
denoiser defines an explicit distribution over categorical label sequences.
At each step, an LCRF conditioned on the input and complete noisy sequence
scores unary and adjacent-label interactions; successive denoising steps
propagate these interactions globally.  Parallel Mean-Field approximates the
required marginals efficiently~\citep{wang-etal-2020-ain}, while categorical
scores support structured decoding.


\section{Model}
\label{sec:model}


\subsection{Standard Sequence Labelling Model}
\label{sec:standard-model}

Given an input sequence $\bm{s}=s_{1}\dots s_{n}$, where \(s_{i}\) is the
\(i^{\text{th}}\) word or character, labelling produces
$\bm{y}=y_{1}\dots y_{n}$, where \(y_{i}\in \mathcal{L}\) labels \(s_{i}\).
Sequence labelling models define a parametrised probability distribution $p_\theta (\bm{y} | \bm{s})$ so labelling amounts to returning the mode $\widehat{\bm{y}} = \argmax_{\bm{y}} p_\theta (\bm{y} | \bm{s})$ and learning parameters $\theta$ is cast as Maximum Likelihood Estimation.
These distributions are written as energy models  $p_\theta (\bm{y} | \bm{s}) \propto \exp f(\bm{s}, \bm{y};\theta)$, computed by a neural network implementing $f$, \ie{} parameters \(\theta\) are the parameters of $f$.
The decomposition of $f$ over sequences is crucial for efficiency.
 
\noindent\textbf{Unigram Models}
sum unary potentials over the sequence $f(\bm{s}, \bm{y};\theta) = \sum_{i=1}^{n} f(\bm{s}, y_i;\theta)$. Typically this is implemented as a Transformer~\cite{vaswani-2017-atten-all-need} 
whose output vector at position $i$ feeds a MLP computing $f$ for all labels at this position.
As a consequence of the decomposition of $f$, $p_{\theta}$ is factorised, \ie{} $p_{\theta}(\bm{y} | \bm{s}) = \prod_{i=1}^{n} p_{\theta}(y_i | \bm{s})$.
This factorisation makes labelling and training efficient but limits fine-grained label interactions.

\noindent\textbf{Bigram Models} sum unary and binary\footnote{These models can be extended to $n$-ary potentials.} potentials over adjacent positions,
$f(\bm{s}, \bm{y};\theta) = \sum_{i=1}^{n} f_1(\bm{s}, y_i;\theta) + \sum_{i=1}^{n-1} f_2(\bm{s}, y_i, y_{i+1};\theta)$.
Transformer outputs feed $f_1$ as in the unigram case.  For $f_2$, an MLP
computes a transition matrix over all pairs $y_i,y_{i+1}$ at each position.
Viterbi decoding~\cite{forney1973viterbi} and Forward/Backward algorithms~\cite{rabiner-1989-tutor-hidden} for marginals make labelling and training linear in sentence length.
LCRFs are difficult to parallelise but approximations such as
 Mean-Field~\cite{wang-etal-2020-ain} or Mean-Regularisation~\cite{corro-etal-2025-bregman} recover the position-wise independent computation and efficiency of unigram models.

\subsection{Discrete Diffusion Models for Labelling}
\label{sec:diffusion-model}

We follow diffusion language models~\cite{hoogeboom-2021-argmax-flows,austin-2021-struc-denois} to define labelling as generation of labels given the input sequence.

\noindent\textbf{Forward Diffusion.}
A sequence of tags \(\bm{y}^{0}\) is altered by a forward diffusion process \(q\) consisting of \(T\) steps \( \bm{y}^{1}\ldots \bm{y}^{T} \) to obtain a random sequence\footnote{All sequences of size  \(|\bm{y}^0|\) are equiprobable 
} \(\bm{y}^{T}\).
Generating such sequences is a Markovian process
\(q(\bm{y}^{1}\ldots \bm{y}^{T} | \bm{y}^{0}) = \prod_{t=1}^T q_{t} (\bm{y}^{t} | \bm{y}^{t-1}) \)
with independent noise at each position \(i\): 
\(q_{t} (\bm{y}^{t} | \bm{y}^{t-1}) = \prod_{i=1}^n q_t(y_i^t | y_{i}^{t-1})\).

Noise distributions are parametrised by a corruption weight \(\beta_{t}\) following a predefined schedule:\footnote{We only consider cosine~\cite{hoogeboom-2021-argmax-flows}.}
\begin{equation*} \label{eq:noise-step}
  q_{t}(y_{i}^{t}| y_{i}^{t-1}) =
  \begin{cases}
    1-\beta_{t} + \frac{\beta_{t}}{|\mathcal{L}|}
      & \text{if } y_{i}^{t} = y_{i}^{t-1},\\
    \frac{\beta_{t}}{|\mathcal{L}|} & \text{otherwise.}
  \end{cases}
\end{equation*}
\noindent Here, $1-\beta_t$ weights label copying and $\beta_t$ weights a
uniform draw, whose per-label probability is $1/|\mathcal{L}|$.
\noindent Each \(q_t(\cdot|\cdot)\) is encoded by a matrix \(Q_t\), allowing \(q_{0|t}\) to be precomputed:
\(q_{0|t} (y_{i}^{t} | y_{i}^{0}) = \sum_{y_{i}^{t-1}}   q_{t}(y_{i}^{t}|y_{i}^{t-1}) q_{0|t-1}(y_{i}^{t-1} | y_{i}^{0})\).

\noindent\textbf{Denoising.}
Our model, following~\citet{hoogeboom-2021-argmax-flows,austin-2021-struc-denois} for language models, produces a parameterised distribution  on label sequences from a random sequence by reversing the diffusion process.
With an abuse of notation, we denote this distribution as $p_{\theta}$.
Reverse generation is also Markovian:
\(p_{\theta} (\bm{y}^0,\bm{y}^{1},\dots,\bm{y}^{T}| \bm{s})
= p(\bm{y}^{T}) \prod_{t=1}^T
p_{\theta}(\bm{y}^{t-1}|\bm{y}^{t}, \bm{s}) \), where the prior
$p(\bm{y}^{T})$ is uniform.
We drop the condition on $\bm{s}$ from notation.

Denoiser \(p_{\theta}\) is implemented by a neural network (\cf{} \S\ref{sec:neural-architecture}).
The same network is used for all \(t\): to add time information, we feed the network with a learned representation of \(t\).
We follow~\citet{ho-2020-denois-diffus} and describe a single denoising step from \(t\) as full denoising followed by \((t-1)\) forward steps:

\vspace{-0.6cm}
\begin{equation*}
\begin{gathered}
  p_{\theta}(\bm{y}^{t-1}|\bm{y}^{t})
  = \sum_{\bm y^{0}} p_{\theta}(\bm{y}^{0}|\bm{y}^{t})
  q(\bm{y}^{t-1} | \bm{y}^{t}, \bm{y}^{0})\\
  \resizebox{\columnwidth}{!}{$\displaystyle
  = \mathbb{E}_{\bm{y}^{0}\sim p_{\theta}(\cdot|\bm{y}^{t})}
  \big[ q(\bm{y}^{t-1} | \bm{y}^{t}, \bm{y}^{0}) \big]
  \;\approx q(\bm{y}^{t-1} | \bm{y}^{t}, \widehat{\bm{y}^{0}} )$}\\
  \text{with } \widehat{\bm{y}^{0}}
  = \mathbb{E}_{\bm{y}^{0}\sim p_{\theta}(\cdot|\bm{y}^{t})}[\bm{y}^{0}].
\end{gathered}
\end{equation*}

A denoising step can be modelled as sampling from $(i)$ the posterior distribution with $(ii)$ the clean sequence $\bm{y}^{0}$ replaced by an expected sequence $\widehat{\bm{y}^{0}}$.
In practice, addressing $(i)$ requires computing the posterior distribution,
expressed with three tractable distributions, from Bayes' theorem and Markovian assumption~:

\vspace{-0.6cm}
\begin{equation*}
\begin{gathered}
  q(\bm{y}^{t-1} | \bm{y}^{t}, \bm{y}^{0})
  = \frac{q( \bm{y}^{t-1}, \bm{y}^{t} | \bm{y}^{0}) }
  {q(\bm{y}^{t}| \bm{y}^{0})} \\
  \resizebox{\columnwidth}{!}{$\displaystyle
  = \frac{q( \bm{y}^{t} | \bm{y}^{t-1} , \bm{y}^{0})
  q( \bm{y}^{t-1} | \bm{y}^{0}) }{q(\bm{y}^{t}| \bm{y}^{0})}
  \;= \frac{q_{t}( \bm{y}^{t} | \bm{y}^{t-1})
  q( \bm{y}^{t-1} | \bm{y}^{0}) }{q(\bm{y}^{t}| \bm{y}^{0})}.$}
\end{gathered}
\end{equation*}

At inference, we skip intermediate timesteps using the \emph{halving} schedule
$t\leftarrow\lfloor t/2\rfloor$.  This reduces denoiser calls from $O(T)$ to
$O(\log T)$, at the possible cost of fewer refinement opportunities.

\begin{figure*}
 \centering
\begin{tikzpicture}[
scale=1.00,
transform shape,
>=latex,
every node/.style={font=\small}
]

\tikzset{
  block/.style={
    draw=SkyBlue!60!black, fill=SkyBlue!20,
    rounded corners, thick,
    minimum width=6.2cm, minimum height=4mm,
    inner sep=1mm, align=center
  },
  mlp/.style={
    draw=Peach!60!black, fill=Peach!20,
    rounded corners, thick,
    minimum width=16mm, minimum height=4mm,
    inner sep=1mm, align=center
  },
  embed/.style={
    draw=Thistle!60!black, fill=Thistle!20,
    rounded corners, thick,
    minimum width=0.6cm, minimum height=4mm,
    inner sep=1mm, align=center
  },
  group/.style={minimum width=6.2cm},
  baselineArea/.style={
    draw=Gray, dashed, rounded corners=4pt,
    fill=Gray!10,
    inner sep=4mm,
  }
}


\node[block] (dit) {Diffusion Transformer};

\node[embed, minimum width=6.2cm, below=1mm of dit] (label_embed) {Label Embeddings};

\node[below=1mm of label_embed] (yti) {\(y^T_i\!\in\! \mathcal{L}\)};
\node[left=1mm of yti] (ytdots1) {\(\ldots\)};
\node[right=1mm of yti] (ytdots2) {\(\ldots\)};
\node[left=1mm of ytdots1] (yt) {\(\bm{y}^T=\)};

\node[right=1mm of ytdots2] (yi_init) {$\sim \mathsf{Cat(\frac{1}{|\mathcal{L}|})}$};

\node[draw, circle, thick, fill=white, above right=0mm of dit, inner sep=1pt, yshift=-2.5mm, xshift=-8mm] (nblocks) {$\times N$};

\node[above=1mm of dit] (zi) {\(\bm{z}_i\!\in\! \mathbb{R}^k\)};
\node[left=1mm of zi] (zdots1) {\(\ldots\)};
\node[right=1mm of zi] (zdots2) {\(\ldots\)};
\node[group, fit=(zdots1)(zi)(zdots2)] (z) {};

\node[mlp, above=3mm of zi, xshift=-12mm] (mlp_l) {Unary MLP};
\node[mlp, above=3mm of zi, xshift=12mm] (mlp_b) {Pairwise MLP};

\draw[-{Latex[length=2mm]}, thick]
  (zi.north) -- (mlp_b.south);
\draw[-{Latex[length=2mm]}, thick]
  (zi.north) -- (mlp_l.south);

\node[above=1mm of mlp_l] (li) {\(\bm{l}_i \!\in\!\mathbb{R}^{|\mathcal{L}|}\)};
\node[above=1mm of mlp_b] (bi) {\(\bm{B}_{i}^{i+\!1}\!\in\!\mathbb{R}^{|\mathcal{L}|\!\times\!|\mathcal{L}|}\)};
\node[left=1mm of li] () {\(\ldots\)};
\node[right=1mm of bi] () {\(\ldots\)};
\node[group, fit=(mlp_l)(mlp_b)] (potentials) {};

\node[block, above=8mm of potentials, draw=SeaGreen!60!black, fill=SeaGreen!20] (crf) {\textbf{LCRF}: Mean-Field Marginal Approx};

\node[above=1mm of crf] (hatyi) {\(\widehat{y^0_i}\!=\mathbb{E}_{p_\theta(\cdot|\bm{y}^t)}\left[y^0_i\right] \!\in[0,\!1]^{|\mathcal{L}|}\)};
\node[left=1mm of hatyi] (hatydots1) {\(\ldots\)};
\node[right=1mm of hatyi] (hatydots2) {\(\ldots\)};

\node[above=4mm of hatyi] (ytm1) {\(\bm{y}^{t-1} \sim q(\cdot|\bm{y}^t,\widehat{\bm{y}^0}) \)};

\draw[-{Latex[length=2mm]}, thick] (hatyi.north) -- node[midway, right] (denoising) {\scriptsize \textit{Sample a label sequence}} (ytm1.south);

\node[embed, left=4mm of label_embed] (t_embed) {T. Emb.};
\node[below=2mm of t_embed] (t) {$t\in\{1,\ldots,T-1\}$};

\node[above left=0mm of t_embed, xshift=10mm] (timestep) {\(\bm{\tau}\!\in\!\mathbb{R}^g\)};

\draw[-{Latex[length=2mm]}, thick, rounded corners=3pt]
($(t_embed.north)+(5mm,0)$) |- ($(dit.west)+(0,-2mm)$);

\draw[-{Latex[length=2mm]}, thick, rounded corners=3pt]
  (ytm1.north) -- ++(0,2mm)
  -| ($(label_embed.east)+(+8mm,0)$) node[pos=0.75, sloped, above] {$t=T,\ldots,1$}
  -- (label_embed.east);

\node[left=61mm of zi] (cdots2) {\(\ldots\)};
\node[left=1mm of cdots2] (ci) {\(\bm{c}_i\!\in\! \mathbb{R}^d\)};
\node[left=1mm of ci] (cdots1) {\(\ldots\)};

\node[block, below=1mm of ci] (w_enc) {Transformer Encoder};
\node[draw, circle, thick, fill=white, above left=0mm of w_enc, inner sep=1pt, yshift=-2.5mm, xshift=8mm] (nblocks) {$\times M$};
\node[embed, minimum width=6.2cm, below=1mm of w_enc] (w_embed) {CharLSTM + Input Embeddings};

\node[below=1mm of w_embed] (wi) {\(s_i\!\in\!\mathcal{V}\)};
\node[left=1mm of wi] (wdots1) {\(\ldots\)};
\node[right=1mm of wi] (wdots2) {\(\ldots\)};
\node[left=1mm of wdots1] (input) {Input \(\bm{s}=\)};

\draw[-{Latex[length=2mm]}, thick, rounded corners=3pt]
  ($(cdots2.east)+(3mm,0)$) -| node[pos=0.38, sloped, below] (context) {\(\bm{C}\in\mathbb{R}^{d\times n}\)} ($(t_embed.north)+(5mm,5mm)$)
  -- (dit.west);

\node[mlp, above=3mm of ci, xshift=-12mm] (base_mlp_l) {Unary MLP};
\node[mlp, above=3mm of ci, xshift=12mm] (base_mlp_b) {Pairwise MLP};

\node[above=1mm of base_mlp_l] (base_li) {\(\bm{l}_i \!\in\!\mathbb{R}^{|\mathcal{L}|}\)};
\node[above=1mm of base_mlp_b] (base_bi) {\(\bm{B}_{i}^{i+\!1}\!\in\!\mathbb{R}^{|\mathcal{L}|\!\times\!|\mathcal{L}|}\)};
\node[left=1mm of base_li] () {\(\ldots\)};
\node[right=1mm of base_bi] () {\(\ldots\)};
\node[group, fit=(base_mlp_l)(base_mlp_b)] (base_potentials) {};

\draw[-{Latex[length=2mm]}, thick]
  (ci.north) -- (base_mlp_b.south);
\draw[-{Latex[length=2mm]}, thick]
  (ci.north) -- (base_mlp_l.south);

\node[block, above=8mm of base_potentials, draw=SeaGreen!60!black, fill=SeaGreen!20] (base_crf) {\textbf{LCRF}: Viterbi};

\node[above=1mm of base_crf] (byi) {\(y_i\!\in\! \mathcal{L}\)};
\node[left=1mm of byi] (bydots1) {\(\ldots\)};
\node[right=1mm of byi] (bydots2) {\(\ldots\)};

\draw[decorate, thick,
  decoration={calligraphic brace, mirror, amplitude=6pt, raise=2pt}
] ([yshift=1pt] bydots2.north east) -- ([yshift=1pt] bydots1.north west);
\node[above=1mm of byi] (by) {\(\bm{\hat{y}}=\argmax_{\bm{y}} P_\theta(\bm{y}|\bm{s})\)};

\begin{scope}[on background layer]
  \node[baselineArea, fit=(base_crf) (base_mlp_l) (base_mlp_b) (by)] (baselineOnly) {};
\end{scope}

\node[anchor=north east, font=\small\itshape]
  at (baselineOnly.north east) {Baseline};

\end{tikzpicture}

 \caption{LCRF (left) and our diffusion denoiser (right), which conditions
 labels on input-unit and timestep embeddings.}
 \label{fig:model}
\end{figure*}
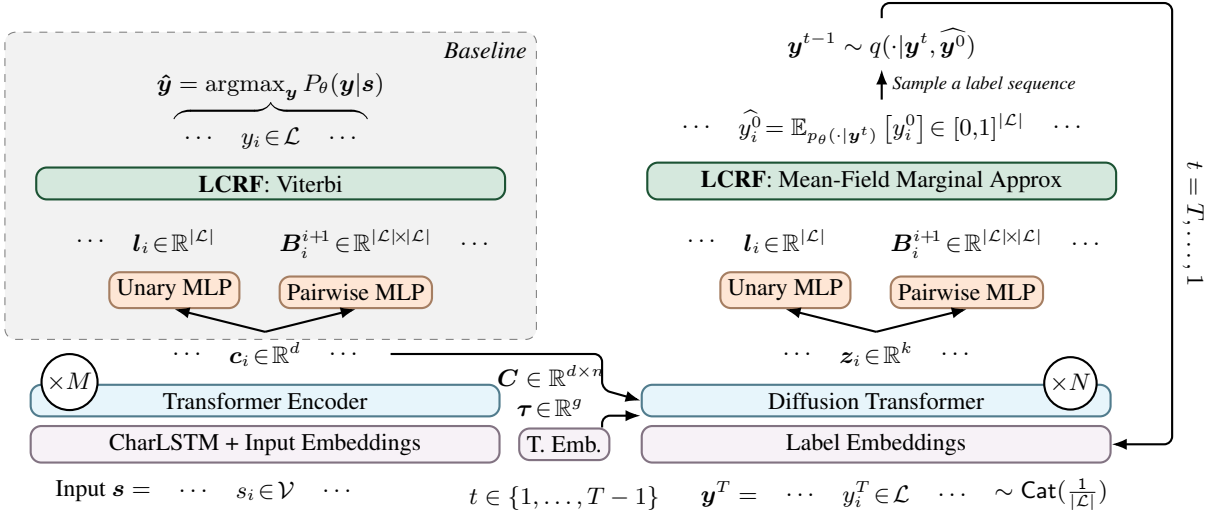

\noindent\textbf{Structured Denoising}
\label{sec:debr-struc}
We can adapt the decoding method to denoisers $p_{\theta}$ implemented by LCRFs.
Remember the denoiser's role is to generate \(\widehat{\bm{y}^{0}}= \mathbb{E}_{\bm{y}^{0}\sim p_{\theta}(\cdot|\bm{y}^{t})}[\bm{y}^{0}]\) the fractional counts or label marginal probabilities.
For LCRFs, exact marginals are tractable in $O(n)$ time using the
Forward/Backward algorithms~\cite{rabiner-1989-tutor-hidden} or backpropagation
through the log-partition~\cite{eisner-2016-inside}.  Repeating these sequential
recurrences inside every denoising call is nevertheless inefficient, and
storing their intermediate states increases training memory.
Instead, we can approximate the LCRF distribution with Mean Regularisation~\cite{corro-etal-2025-bregman} or Mean-Field~\cite{wang-etal-2020-ain}.
We use the latter and show we can exploit structures with diffusion models while remaining efficient.

\textbf{Training}
minimises a sampled variational objective augmented with direct denoising
supervision.  Its variational component follows the lower bound:
\vspace{-0.6cm}
\begin{align*}
&\log p_{\theta}(\bm y^{0})
 = \log \sum_{\bm y^{1},\dots,\bm y^{T}}
 p_{\theta}(\bm{y}^{0},\bm{y}^{1},\dots,\bm{y}^{T}) \\
&\ge \E_{\bm y^{1}\sim q_{0|1}(\cdot\mid \bm y^{0})}
   \Big[\log p_{\theta}(\bm y^{0}\mid \bm y^{1})\Big] \nonumber\\
&-\sum_{t=2}^{T}
  \E_{\bm y^{t}\sim q_{0|t}(\cdot\mid \bm y^{0})}
  \left[\,\KLsplit{q(\bm y^{t-1}\mid \bm y^{t},\bm y^{0})}
                 {p_{\theta}(\bm y^{t-1}\mid \bm y^{t})}\,\right] \nonumber\\
&- \KL\Big(q_{0|T}(\bm y^{T}\mid \bm y^{0}) \,\big\|\, p(\bm y^{T})\Big),
\end{align*}
\noindent{}where \(D_{KL}\) is the Kullback--Leibler divergence.
The last term is ignored since both distributions are uniform and their divergence is zero.
During training, we sample one shared $t\sim\mathcal U\{1,\ldots,T-1\}$ per
sequence, draw $\bm y^t\sim q_{0|t}(\cdot\mid\bm y^0)$, and supply its timestep
embedding at every position.

For $t\geq2$, the sampled $\bm y^t$ estimates only the corresponding KL term:
\begin{align*}
   &D_{KL}[q(\bm{y}^{t-1} | \bm{y}^{t}, \bm{y}^{0})||p_{\theta}(\bm{y}^{t-1}|\bm{y}^{t})] \\
  &\approx D_{KL}[q(\bm{y}^{t-1} | \bm{y}^{t}, \bm{y}^{0})||q(\bm{y}^{t-1} | \bm{y}^{t}, \widehat{\bm{y}^{0}})],
\end{align*}
\noindent{}At $t=1$, the gold posterior is a point mass, so this KL reduces to
reconstruction.  We call the sampled term $\mathcal L_{\mathrm{post}}$ and add
direct token supervision
$\mathcal L_{\mathrm{den}}=-|\mathcal I|^{-1}\sum_{i\in\mathcal I}
\log \widehat y_i^0[y_i^0]$, where $\mathcal I$ contains non-padding positions.
For the unigram and Mean-Field denoisers, the resulting hybrid loss is
$\mathcal L_{\mathrm{hyb}}=\mathcal L_{\mathrm{post}}+
\mathcal L_{\mathrm{den}}$~\cite{austin-2021-struc-denois}.

\subsection{Neural Architecture}
\label{sec:neural-architecture}

The architecture depicted in Fig.\ref{fig:model} implements potential functions, $f$ for unigrams or $f_1,f_2$ for bigrams, as defined in  §\ref{sec:standard-model}, that are used to define probabilities.
All models convert input units $s_1,\dots,s_n$ to non-contextual representations with a look-up table and a charLSTM~\citep{lample-etal-2016-neural}.
These are contextualised with
Transformers~\cite{vaswani-2017-atten-all-need} to obtain vectors
$e_1,\dots,e_n$.  We call this task-trained combination our \emph{scratch
encoder}; alternatively, it can be replaced with a pretrained
model such as BERT or RoBERTa.
The unigram model is parametrised by $n$ vectors $l_i$ of $|\mathcal{L}|$ scores computed by a MLP from $e_i$. 
The bigram model adds $|\mathcal{L}|\times|\mathcal{L}|$ scores for label transitions from one position to the next.
These are computed for each position by a MLP.
 
Our diffusion denoisers are based on Diffusion Transformer blocks~\citep{Peebles2022DiT}.
They take a sequence of label embeddings from a trainable look-up table,
corresponding to noisy labels and, as context, the concatenation of input
embeddings and a trainable timestep embedding.
After these blocks an MLP converts position vectors to unigram, and possibly bigram, scores.



\section{Experiments}
\label{sec:experiments}

\begin{table*}
\centering
\small
\setlength{\tabcolsep}{2.6pt}
\begin{tabular}{llrrrrrrrrrrr}
\toprule
& & \multicolumn{5}{c}{POS accuracy}
& \multicolumn{3}{c}{NER}
& \multicolumn{3}{c}{Chinese Seg+POS} \\
\cmidrule(lr){3-7}\cmidrule(lr){8-10}\cmidrule(l){11-13}
Encoder & Model & EN & DE & FR & NL & Avg.
& P & R & F1
& Seg.\ F1 & XPOS & Joint F1 \\
\midrule
\multirow{5}{*}{Scratch} & \sys{Unigram}
  & 91.48 & 92.38 & 94.58 & 91.94 & 92.60
  & 74.89 & 72.70 & 73.75
  & 86.25 & 83.17 & 77.01 \\
& \sys{Mean-Field}
  & 93.51 & 93.88 & 96.02 & 93.84 & 94.31
  & 77.73 & 74.93 & 76.31
  & 90.84 & 87.14 & 83.34 \\
& \sys{LCRF}
  & \underline{94.02} & 94.11 & 96.60 & \underline{94.00} & \underline{94.68}
  & 80.02 & \underline{77.80} & 78.89
  & \underline{91.09} & \underline{87.17} & 83.78 \\
& \sys{DiffusionSL}
  & 93.83 & \underline{94.29} & \underline{96.67} & 93.49 & 94.57
  & \textbf{82.34} & 77.48 & \underline{79.84}
  & 90.37 & 86.94 & \underline{83.89} \\
& \textbf{\sys{Diffusion-MF}}
  & \textbf{95.06} & \textbf{94.85} & \textbf{97.39} & \textbf{94.93} & \textbf{95.56}
  & \underline{81.46} & \textbf{79.86} & \textbf{80.65}
  & \textbf{92.89} & \textbf{89.50} & \textbf{86.69} \\
\midrule
\multirow{5}{*}{Pretrained} & \sys{Unigram}
  & \underline{98.41} & \textbf{96.91} & \underline{98.26} & \underline{97.56} & \underline{97.78}
  & 91.24 & 92.25 & 91.74
  & 98.03 & 96.41 & 95.25 \\
& \sys{Mean-Field}
  & 98.28 & 96.80 & 98.14 & 97.35 & 97.64
  & 91.74 & 92.45 & 92.09
  & 98.06 & 96.43 & 95.29 \\
& \sys{LCRF}
  & \underline{98.41} & 96.85 & \underline{98.26} & 97.48 & 97.75
  & 91.71 & \underline{92.50} & 92.11
  & 98.02 & 96.42 & 95.26 \\
& \sys{DiffusionSL}
  & 98.30 & 96.77 & \underline{98.26} & 97.44 & 97.69
  & \textbf{92.58} & 92.18 & \underline{92.38}
  & \underline{98.09} & \underline{96.62} & \underline{95.59} \\
& \textbf{\sys{Diffusion-MF}}
  & \textbf{98.50} & \underline{96.89} & \textbf{98.30} & \textbf{97.62} & \textbf{97.83}
  & \underline{92.27} & \textbf{92.75} & \textbf{92.51}
  & \textbf{98.27} & \textbf{96.77} & \textbf{95.71} \\
\bottomrule
\end{tabular}
\caption{Test results (4-seed means) with dev-selected decoding. Bold and
underlining mark the best and runner-up per encoder; metrics are defined in
\S\ref{sec:experimental-setup}.}
\label{tab:main-results}
\end{table*}


\subsection{Experimental Setup}
\label{sec:experimental-setup}

\paragraph{Tasks and metrics.}
The tasks vary in output granularity, label set, and evaluation.  Multilingual
POS assigns 16--17 UPOS tags to words in four UD~2.15 treebanks
\citep{11234/1-5787} and is evaluated by token accuracy.  CoNLL-2003 English
NER~\citep{tjong-kim-sang-de-meulder-2003-introduction} uses 17 BMES word
labels (four entity types plus \texttt{O}); we convert BIO annotations to BMES
and report entity precision, recall, and F1.  Chinese GSDSimp assigns one of
104 observed boundary--XPOS combinations to each character (four boundary
states and 41 XPOS values).  Its primary metric is joint F1, which requires
both word span and XPOS tag to match; we also report segmentation F1 and XPOS
accuracy on gold words.  Appendix~\ref{sec:dataset-statistics} gives dataset
statistics and metric definitions.

\paragraph{Structure and validity.}
Adjacent-label transitions express affinities in every task, but only NER and
Chinese impose hard well-formedness constraints.  Every POS label sequence is
valid.  NER and Chinese BMES outputs must open and close complete spans or
words and keep the entity type or XPOS tag consistent within each object.  A
valid sequence maps directly to evaluated objects; for a malformed sequence,
the evaluator retains complete spans and singletons and ignores malformed
fragments.

\paragraph{Systems and decoding.}
We compare \sys{Unigram}, \sys{Mean-Field}, and \sys{LCRF}, their discrete
diffusion counterparts, and continuous \sys{DiffusionSL}
\citep{huang-etal-2023-diffusionsl}; \sys{Diffusion-MF} is our main model.
Model names describe the output distribution and internal inference, whereas
decoder names describe only the final prediction.  Unigram families naturally
return categorical unary scores, while Mean-Field families return approximate
LCRF marginals.  Both support position-wise argmax and, for constrained tasks,
\emph{Mask-Viterbi}, which adds fixed legality constraints but no learned
transition preferences.  LCRF families also expose learned transition scores
and use Viterbi with the same task constraints.  \sys{DiffusionSL} instead
returns continuous bit vectors and natively thresholds each position, so it
cannot rank label alternatives.  We implement a distance-based unary extension
that enables Mask-Viterbi (Appendix~\ref{sec:diffusionsl-mask-viterbi}).  Thus
POS uses argmax for Unigram and Mean-Field families, Viterbi for LCRFs, and
native thresholding for \sys{DiffusionSL}.  On NER and Chinese, we additionally
test Mask-Viterbi wherever unary scores are available.  Development performance
selects each system's configuration and decoder;
Appendices~\ref{sec:experimental-details}--\ref{sec:full-results} give full
evaluation and seed details.

\subsection{Results and Analysis}
\label{sec:overall-results}
\label{sec:diffusion-mf-analysis}
\label{sec:main-efficiency}

\paragraph{Overall results.}
Table~\ref{tab:main-results} shows that \sys{Diffusion-MF} leads every task
under both encoder regimes and outperforms \sys{DiffusionSL} throughout.  Its
larger advantage with scratch encoders suggests that structured categorical
diffusion is most useful when representations are learned only from task data.

\begin{table}
\centering
\small
\setlength{\tabcolsep}{3pt}
\begin{tabular}{@{}lccc@{}}
\toprule
Setting & \sys{Diff.-Uni} & \sys{Diff.-LCRF} & \sys{Diff.-MF} \\
\midrule
POS scratch & 95.45 & 92.34 & \textbf{95.56} \\
POS pretrained & 97.82 & 97.77 & \textbf{97.83} \\
NER scratch & 80.14 & 79.83 & \textbf{80.65} \\
NER pretrained & 92.49 & 92.43 & \textbf{92.51} \\
Seg+POS scratch & 86.52 & 84.20 & \textbf{86.69} \\
Seg+POS pretrained & 95.68 & 95.38 & \textbf{95.71} \\
\bottomrule
\end{tabular}
\caption{Primary-metric denoiser comparison after dev selection (4-seed means).}
\label{tab:diffusion-comparison}
\end{table}


\paragraph{Structured denoiser.}
\label{sec:denoiser-comparison}

Within categorical diffusion, \sys{Diffusion-Uni} already performs strongly,
showing that iterative global refinement supplies much of the improvement over
non-diffusion baselines.  Mean-Field nevertheless improves it in all six
settings, indicating that adjacent-label structure adds consistent value.
\sys{Diffusion-LCRF} never improves on Mean-Field and is weakest in scratch
settings.  Because it uses a different structured objective, this system-level
comparison does not isolate exact versus approximate inference.

\begin{table}[!htbp]
\centering
\small
\setlength{\tabcolsep}{4pt}
\begin{tabular}{@{}lrrr@{}}
\toprule
Setting & POS acc. & NER F1 & Chinese joint F1 \\
\midrule
\multicolumn{4}{@{}l}{\textit{Training objective}} \\
Full hybrid & $\mathbf{95.02{\pm}.11}$ & $\mathbf{80.92{\pm}.70}$ & $\mathbf{86.59{\pm}.35}$ \\
$-\mathcal L_{\rm den}$ & $94.27{\pm}.17$ & $79.69{\pm}.69$ & $85.97{\pm}.17$ \\
$-\mathcal L_{\rm post}$ & $94.96{\pm}.12$ & $80.56{\pm}.63$ & $86.45{\pm}.28$ \\
\midrule
\multicolumn{4}{@{}l}{\textit{Denoiser calls}} \\
5 & $\mathbf{95.05{\pm}.09}$ & $\mathbf{81.58{\pm}.69}$ & $86.42{\pm}.38$ \\
10 & $95.02{\pm}.11$ & $80.92{\pm}.70$ & $\mathbf{86.59{\pm}.35}$ \\
20 & $95.00{\pm}.14$ & $81.47{\pm}.51$ & $86.50{\pm}.28$ \\
\bottomrule
\end{tabular}
\caption{Test results for diffusion-specific ablations of scratch
\sys{Diffusion-MF}, averaged over three seeds.  The objective
ablation removes either direct clean-label denoising supervision
($\mathcal L_{\rm den}$) or posterior matching ($\mathcal L_{\rm post}$).
Ten calls reproduce the base-2 halving schedule used in the main experiments.}
\label{tab:diffusion-component-ablation}
\end{table}

\paragraph{Objective and inference budget.}
\label{sec:main-diffusion-components}
Table~\ref{tab:diffusion-component-ablation} reports controlled scratch-model
comparisons over three matched seeds.  Removing either loss weakens all three
tasks, with direct clean-label supervision having the larger and more
consistent effect.  Varying the number of denoiser calls yields no monotonic
improvement, so additional refinement alone does not explain the main gains
(Appendix~\ref{sec:diffusion-component-ablation}).

\paragraph{Efficiency.}

\begin{table}
\centering
\small
\setlength{\tabcolsep}{3pt}
\begin{tabular}{lrrrrrr}
\toprule
& \multicolumn{3}{c}{Training} & \multicolumn{3}{c}{Inference} \\
\cmidrule(lr){2-4}\cmidrule(l){5-7}
Task & \sys{Uni} & \sys{MF} & \sys{LCRF}
     & \sys{Uni} & \sys{MF} & \sys{LCRF} \\
\midrule
POS (EWT) & 267.9 & 225.9 & 82.3 & 337.1 & 314.0 & 253.4 \\
NER & 307.0 & 237.1 & 93.6 & 345.9 & 273.7 & 56.8 \\
Seg.+POS & 123.2 & 85.9 & 27.5 & 93.2 & 42.3 & 17.3 \\
\bottomrule
\end{tabular}
\caption{Median sentences/s for the three diffusion denoisers on one A100.}
\label{tab:efficiency}
\end{table}


Table~\ref{tab:efficiency} isolates denoiser cost under matched conditions.
Mean-Field is consistently faster than exact LCRF during training and
inference.  Its overhead relative to \sys{Diffusion-Uni} grows with the label
set---modest for POS and largest for Chinese---as expected for pairwise
updates, but remains markedly cheaper than exact LCRF
(Appendix~\ref{sec:additional-analyses}).


\section{Conclusion}
\label{sec:conclusion}
We introduced \sys{Diffusion-MF}, which approximates LCRF marginals with
parallel Mean-Field inference at every discrete denoising step.  It achieves
the best primary score across all six task--encoder settings while remaining
faster than exact LCRF denoising.  These results show that discrete diffusion
can combine global iterative refinement with explicit sequence structure.


\section*{Limitations}
\label{sec:limitations}

Our evidence is limited to linear-chain labelling on multilingual POS, English
NER, and Chinese segmentation--POS. Although these tasks vary in label-set size
and output constraints, they do not establish how the method behaves with
higher-order structures, longer outputs, or other structured-prediction tasks.
Moreover, the small differences between pretrained systems should be
interpreted cautiously despite averaging over four seeds.

Diffusion inference still requires repeated denoiser calls, and
\sys{Diffusion-MF} additionally performs several Mean-Field iterations. It is
therefore slower than unstructured diffusion, especially for large label sets.
Parallel Mean-Field is approximate and has no general convergence guarantee;
its accuracy can depend on the number of iterations and the strength of the
pairwise potentials. The \sys{Diffusion-LCRF} comparison reflects
differences in training objectives, so its scores do not establish that
exact structured denoising is inferior.

\section*{Ethical Considerations}
\label{sec:ethic-cons}

We believe that our work does not raise ethical concerns.
We present a novel architecture for sequence labelling based on diffusion and structured prediction and we test it on publicly available data.

We acknowledge the environmental impact of the energy cost of training neural models.

\bibliography{main}

\begin{thebibliography}{37}
\providecommand{\natexlab}[1]{#1}

\bibitem[{Austin et~al.(2021)Austin, Johnson, Ho, Tarlow, and van~den
  Berg}]{austin-2021-struc-denois}
Jacob Austin, Daniel~D. Johnson, Jonathan Ho, Daniel Tarlow, and Rianne van~den
  Berg. 2021.
\newblock \href
  {https://proceedings.neurips.cc/paper_files/paper/2021/file/958c530554f78bcd8e97125b70e6973d-Paper.pdf}
  {Structured denoising diffusion models in discrete state-spaces}.
\newblock In \emph{Advances in Neural Information Processing Systems},
  volume~34, pages 17981--17993. Curran Associates, Inc.

\bibitem[{Chang et~al.(2022)Chang, Zhang, Jiang, Liu, and
  Freeman}]{Chang2022MaskGITMG}
Huiwen Chang, Han Zhang, Lu~Jiang, Ce~Liu, and William~T. Freeman. 2022.
\newblock \href {https://api.semanticscholar.org/CorpusID:246680316} {Maskgit:
  Masked generative image transformer}.
\newblock \emph{2022 IEEE/CVF Conference on Computer Vision and Pattern
  Recognition (CVPR)}, pages 11305--11315.

\bibitem[{Chen et~al.(2022)Chen, Zhang, and Hinton}]{Chen2022AnalogBG}
Ting Chen, Ruixiang Zhang, and Geoffrey~E. Hinton. 2022.
\newblock \href {https://api.semanticscholar.org/CorpusID:251402961} {Analog
  bits: Generating discrete data using diffusion models with
  self-conditioning}.
\newblock \emph{ArXiv}, abs/2208.04202.

\bibitem[{Corro et~al.(2025)Corro, Lacroix, and Roux}]{corro-etal-2025-bregman}
Caio Corro, Mathieu Lacroix, and Joseph~Le Roux. 2025.
\newblock \href {https://aclanthology.org/2025.acl-long.1430/} {Bregman
  conditional random fields: Sequence labeling with parallelizable inference
  algorithms}.
\newblock In \emph{Proceedings of the 63rd Annual Meeting of the Association
  for Computational Linguistics (Volume 1: Long Papers)}, pages 29557--29574,
  Vienna, Austria. Association for Computational Linguistics.

\bibitem[{Cui et~al.(2021)Cui, Che, Liu, Qin, Wang, and
  Hu}]{cui-etal-2021-whole-word-masking}
Yiming Cui, Wanxiang Che, Ting Liu, Bing Qin, Shijin Wang, and Guoping Hu.
  2021.
\newblock \href {https://doi.org/10.1109/TASLP.2021.3124365} {Pre-training with
  whole word masking for chinese {BERT}}.
\newblock \emph{{IEEE/ACM} Transactions on Audio, Speech, and Language
  Processing}, 29:3504--3514.

\bibitem[{Devlin et~al.(2019)Devlin, Chang, Lee, and
  Toutanova}]{devlin-etal-2019-bert}
Jacob Devlin, Ming-Wei Chang, Kenton Lee, and Kristina Toutanova. 2019.
\newblock \href {https://doi.org/10.18653/v1/N19-1423} {{BERT}: Pre-training of
  deep bidirectional transformers for language understanding}.
\newblock In \emph{Proceedings of the 2019 Conference of the North American
  Chapter of the Association for Computational Linguistics: Human Language
  Technologies, Volume 1 (Long and Short Papers)}, pages 4171--4186,
  Minneapolis, Minnesota. Association for Computational Linguistics.

\bibitem[{Domke(2012)}]{Domke2012}
Justin Domke. 2012.
\newblock \href {https://proceedings.mlr.press/v22/domke12.html} {Generic
  methods for optimization-based modeling}.
\newblock In \emph{Proceedings of the Fifteenth International Conference on
  Artificial Intelligence and Statistics}, volume~22 of \emph{Proceedings of
  Machine Learning Research}, pages 318--326, La Palma, Canary Islands. PMLR.

\bibitem[{Eisner(2016)}]{eisner-2016-inside}
Jason Eisner. 2016.
\newblock \href {https://doi.org/10.18653/v1/W16-5901} {Inside-outside and
  forward-backward algorithms are just backprop (tutorial paper)}.
\newblock In \emph{Proceedings of the Workshop on Structured Prediction for
  {NLP}}, pages 1--17, Austin, TX. Association for Computational Linguistics.

\bibitem[{Forney(1973)}]{forney1973viterbi}
George~David Forney. 1973.
\newblock \href {https://doi.org/10.1109/PROC.1973.9030} {The {V}iterbi
  algorithm}.
\newblock \emph{Proceedings of the IEEE}, 61(3):268--278.

\bibitem[{Gong et~al.(2022)Gong, Li, Feng, Wu, and Kong}]{Gong2022DiffuSeqST}
Shansan Gong, Mukai Li, Jiangtao Feng, Zhiyong Wu, and Lingpeng Kong. 2022.
\newblock \href {https://api.semanticscholar.org/CorpusID:252917661} {Diffuseq:
  Sequence to sequence text generation with diffusion models}.
\newblock \emph{ArXiv}, abs/2210.08933.

\bibitem[{Gu et~al.(2021)Gu, Chen, Bao, Wen, Zhang, Chen, Yuan, and
  Guo}]{Gu2021VectorQD}
Shuyang Gu, Dong Chen, Jianmin Bao, Fang Wen, Bo~Zhang, Dongdong Chen, Lu~Yuan,
  and Baining Guo. 2021.
\newblock \href {https://api.semanticscholar.org/CorpusID:244714856} {Vector
  quantized diffusion model for text-to-image synthesis}.
\newblock \emph{2022 IEEE/CVF Conference on Computer Vision and Pattern
  Recognition (CVPR)}, pages 10686--10696.

\bibitem[{Hershey et~al.(2014)Hershey, Roux, and
  Weninger}]{Hershey2014DeepUnfolding}
John~R. Hershey, Jonathan~Le Roux, and Felix Weninger. 2014.
\newblock \href {https://arxiv.org/abs/1409.2574} {Deep unfolding: Model-based
  inspiration of novel deep architectures}.
\newblock \emph{ArXiv}, abs/1409.2574.

\bibitem[{Ho et~al.(2020)Ho, Jain, and Abbeel}]{ho-2020-denois-diffus}
Jonathan Ho, Ajay Jain, and Pieter Abbeel. 2020.
\newblock \href
  {https://proceedings.neurips.cc/paper_files/paper/2020/file/4c5bcfec8584af0d967f1ab10179ca4b-Paper.pdf}
  {Denoising diffusion probabilistic models}.
\newblock In \emph{Advances in Neural Information Processing Systems},
  volume~33, pages 6840--6851. Curran Associates, Inc.

\bibitem[{Hoogeboom et~al.(2021)Hoogeboom, Nielsen, Jaini, Forr\'{e}, and
  Welling}]{hoogeboom-2021-argmax-flows}
Emiel Hoogeboom, Didrik Nielsen, Priyank Jaini, Patrick Forr\'{e}, and Max
  Welling. 2021.
\newblock \href
  {https://proceedings.neurips.cc/paper_files/paper/2021/file/67d96d458abdef21792e6d8e590244e7-Paper.pdf}
  {Argmax flows and multinomial diffusion: Learning categorical distributions}.
\newblock In \emph{Advances in Neural Information Processing Systems},
  volume~34, pages 12454--12465. Curran Associates, Inc.

\bibitem[{Huang et~al.(2023{\natexlab{a}})Huang, Cao, Zhao, and
  Liu}]{huang-etal-2023-diffusionsl}
Ziyang Huang, Pengfei Cao, Jun Zhao, and Kang Liu. 2023{\natexlab{a}}.
\newblock \href {https://doi.org/10.18653/v1/2023.findings-emnlp.860}
  {{D}iffusion{SL}: Sequence labeling via tag diffusion process}.
\newblock In \emph{Findings of the Association for Computational Linguistics:
  EMNLP 2023}, pages 12902--12920, Singapore. Association for Computational
  Linguistics.

\bibitem[{Huang et~al.(2023{\natexlab{b}})Huang, Cao, Zhao, and
  Liu}]{Huang2023}
Ziyang Huang, Pengfei Cao, Jun Zhao, and Kang Liu. 2023{\natexlab{b}}.
\newblock \href {https://arxiv.org/abs/2309.08006} {Diffusionsl: Sequence
  labeling via tag diffusion process}.
\newblock In \emph{Proceedings of the 2023 Conference on Empirical Methods in
  Natural Language Processing}.
\newblock To appear.

\bibitem[{Jayasumana et~al.(2024)Jayasumana, Glasner, Ramalingam, Veit,
  Chakrabarti, and Kumar}]{jayasumana-2024-markov}
Sadeep Jayasumana, Daniel Glasner, Srikumar Ramalingam, Andreas Veit, Ayan
  Chakrabarti, and Sanjiv Kumar. 2024.
\newblock Markovgen: Structured prediction for efficient text-to-image
  generation.
\newblock In \emph{Proceedings of the IEEE/CVF Conference on Computer Vision
  and Pattern Recognition (CVPR)}, pages 9316--9325.

\bibitem[{Lafferty et~al.(2001)Lafferty, McCallum, and
  Pereira}]{lafferty-2001-condit-random-field}
John~D. Lafferty, Andrew McCallum, and Fernando C.~N. Pereira. 2001.
\newblock Conditional random fields: Probabilistic models for segmenting and
  labeling sequence data.
\newblock In \emph{Proceedings of the Eighteenth International Conference on
  Machine Learning {(ICML} 2001), Williams College, Williamstown, MA, USA, June
  28 - July 1, 2001}, pages 282--289. Morgan Kaufmann.

\bibitem[{Lample et~al.(2016)Lample, Ballesteros, Subramanian, Kawakami, and
  Dyer}]{lample-etal-2016-neural}
Guillaume Lample, Miguel Ballesteros, Sandeep Subramanian, Kazuya Kawakami, and
  Chris Dyer. 2016.
\newblock \href {https://doi.org/10.18653/v1/N16-1030} {Neural architectures
  for named entity recognition}.
\newblock In \emph{Proceedings of the 2016 Conference of the North {A}merican
  Chapter of the Association for Computational Linguistics: Human Language
  Technologies}, pages 260--270, San Diego, California. Association for
  Computational Linguistics.

\bibitem[{Li et~al.(2022)Li, Thickstun, Gulrajani, Liang, and
  Hashimoto}]{Li2022DiffusionLMIC}
Xiang~Lisa Li, John Thickstun, Ishaan Gulrajani, Percy Liang, and Tatsunori
  Hashimoto. 2022.
\newblock \href {https://api.semanticscholar.org/CorpusID:249192356}
  {Diffusion-lm improves controllable text generation}.
\newblock \emph{ArXiv}, abs/2205.14217.

\bibitem[{Liu et~al.(2019)Liu, Ott, Goyal, Du, Joshi, Chen, Levy, Lewis,
  Zettlemoyer, and Stoyanov}]{DBLP:journals/corr/abs-1907-11692}
Yinhan Liu, Myle Ott, Naman Goyal, Jingfei Du, Mandar Joshi, Danqi Chen, Omer
  Levy, Mike Lewis, Luke Zettlemoyer, and Veselin Stoyanov. 2019.
\newblock \href {https://arxiv.org/abs/1907.11692} {Roberta: {A} robustly
  optimized {BERT} pretraining approach}.
\newblock \emph{CoRR}, abs/1907.11692.

\bibitem[{Murphy et~al.(1999)Murphy, Weiss, and Jordan}]{Murphy1999Loopy}
Kevin~P. Murphy, Yair Weiss, and Michael~I. Jordan. 1999.
\newblock Loopy belief propagation for approximate inference: An empirical
  study.
\newblock In \emph{Proceedings of the Fifteenth Conference on Uncertainty in
  Artificial Intelligence}, pages 467--475. Morgan Kaufmann.

\bibitem[{Peebles and Xie(2022)}]{Peebles2022DiT}
William Peebles and Saining Xie. 2022.
\newblock \href {https://arxiv.org/abs/2212.09748} {Scalable diffusion models
  with transformers}.
\newblock \emph{arXiv preprint arXiv:2212.09748}.

\bibitem[{Qiu et~al.(2025)Qiu, Dong, Zhang, Xing, and
  Huang}]{Qiu2025DiffusionCRF}
Yunfei Qiu, Libo Dong, Wenwen Zhang, Haoran Xing, and Junwei Huang. 2025.
\newblock \href {https://doi.org/10.1038/s41598-025-04036-x} {A diffusion
  enhanced crf and bilstm framework for accurate entity recognition}.
\newblock \emph{Scientific Reports}, 15:19670.

\bibitem[{Rabiner(1989)}]{rabiner-1989-tutor-hidden}
Lawrence~R. Rabiner. 1989.
\newblock A tutorial on hidden {Markov} models and selected applications in
  speech recognition.
\newblock \emph{Proceedings of the IEEE}, 77(2):257--285.

\bibitem[{Ranasinghe et~al.(2024)Ranasinghe, Jayasumana, Veit, Chakrabarti,
  Glasner, Ryoo, Ramalingam, and Kumar}]{Ranasinghe2024LatentCRF}
Kanchana Ranasinghe, Sadeep Jayasumana, Andreas Veit, Ayan Chakrabarti, Daniel
  Glasner, Michael~S Ryoo, Srikumar Ramalingam, and Sanjiv Kumar. 2024.
\newblock \href {https://arxiv.org/abs/2412.18596} {Latentcrf: Continuous crf
  for efficient latent diffusion}.
\newblock \emph{Preprint}, arXiv:2412.18596.

\bibitem[{Shen et~al.(2023)Shen, Song, Tan, Li, Lu, and Zhuang}]{Shen2023}
Yongliang Shen, Kaitao Song, Xu~Tan, Dongsheng Li, Weiming Lu, and Yueting
  Zhuang. 2023.
\newblock \href {https://aclanthology.org/2023.acl-long.214} {Diffusionner:
  Boundary diffusion for named entity recognition}.
\newblock In \emph{Proceedings of the 61st Annual Meeting of the Association
  for Computational Linguistics (Volume 1: Long Papers)}, pages 3875--3890.
  Association for Computational Linguistics.

\bibitem[{Stoyanov and Eisner(2011)}]{Stoyanov2011ApproxInferencePolicies}
Veselin Stoyanov and Jason Eisner. 2011.
\newblock \href {https://www.cs.jhu.edu/~jason/} {Learning cost-aware,
  loss-aware approximate inference policies for probabilistic graphical
  models}.
\newblock In \emph{NIPS Workshop on Structured Prediction and Approximate
  Inference}.
\newblock Workshop on Advances in Structured Prediction.

\bibitem[{Sun et~al.(2023)Sun, Yu, Dai, Schuurmans, and
  Dai}]{Sun2023CategoricalDiffusion}
Haoran Sun, Lijun Yu, Bo~Dai, Dale Schuurmans, and Hanjun Dai. 2023.
\newblock \href {https://openreview.net/forum?id=BYWWwSY2G5s} {Score-based
  continuous-time discrete diffusion models}.
\newblock In \emph{The Eleventh International Conference on Learning
  Representations}.

\bibitem[{Tjong Kim~Sang and
  De~Meulder(2003)}]{tjong-kim-sang-de-meulder-2003-introduction}
Erik~F. Tjong Kim~Sang and Fien De~Meulder. 2003.
\newblock \href {https://aclanthology.org/W03-0419/} {Introduction to the
  {C}o{NLL}-2003 shared task: Language-independent named entity recognition}.
\newblock In \emph{Proceedings of the Seventh Conference on Natural Language
  Learning at {HLT}-{NAACL} 2003}, pages 142--147.

\bibitem[{Vaswani et~al.(2017)Vaswani, Shazeer, Parmar, Uszkoreit, Jones,
  Gomez, Kaiser, and Polosukhin}]{vaswani-2017-atten-all-need}
Ashish Vaswani, Noam Shazeer, Niki Parmar, Jakob Uszkoreit, Llion Jones,
  Aidan~N Gomez, \L~ukasz Kaiser, and Illia Polosukhin. 2017.
\newblock \href
  {https://proceedings.neurips.cc/paper_files/paper/2017/file/3f5ee243547dee91fbd053c1c4a845aa-Paper.pdf}
  {Attention is all you need}.
\newblock In \emph{Advances in Neural Information Processing Systems},
  volume~30. Curran Associates, Inc.

\bibitem[{Wainwright and Jordan(2008)}]{Wainwright2008}
Martin~J. Wainwright and Michael~I. Jordan. 2008.
\newblock \href {https://doi.org/10.1561/2200000001} {Graphical models,
  exponential families, and variational inference}.
\newblock \emph{Foundations and Trends in Machine Learning}, 1(1--2):1--305.

\bibitem[{Wang et~al.(2020)Wang, Jiang, Bach, Wang, Huang, Huang, and
  Tu}]{wang-etal-2020-ain}
Xinyu Wang, Yong Jiang, Nguyen Bach, Tao Wang, Zhongqiang Huang, Fei Huang, and
  Kewei Tu. 2020.
\newblock \href {https://doi.org/10.18653/v1/2020.emnlp-main.485} {{AIN}: Fast
  and accurate sequence labeling with approximate inference network}.
\newblock In \emph{Proceedings of the 2020 Conference on Empirical Methods in
  Natural Language Processing (EMNLP)}, pages 6019--6026, Online. Association
  for Computational Linguistics.

\bibitem[{Yedidia et~al.(2005)Yedidia, Freeman, and Weiss}]{Yedidia2005}
Jonathan~S. Yedidia, William~T. Freeman, and Yair Weiss. 2005.
\newblock Constructing free-energy approximations and generalized belief
  propagation algorithms.
\newblock \emph{IEEE Transactions on Information Theory}, 51(7):2282--2312.

\bibitem[{Zeman et~al.(2024)Zeman, Nivre, Abrams, Ackermann, Aepli, Aghaei,
  Agi{\'c}, Ahmadi, Ahrenberg, Ajede, Akhundjanova, Akkurt, Aleksandravi{\v
  c}i{\=u}t{\.e}, Alfina, Algom, Alnajjar, Alzetta, Andersen, Andrews,
  Antonsen, Aoyama, Aplonova, Aquino, Aragon, Aranes, Aranzabe, Ar{\i}can,
  Arnard{\'o}ttir, Arutie, Arwidarasti, Asahara, {\'A}sgeirsd{\'o}ttir, Aslan,
  Asmazo{\u g}lu, Ateyah, Atmaca, Attia, Atutxa, Augustinus, Avel{\~a}s,
  Badmaeva, Balasubramani, Ballesteros, Banerjee, Bank, Barbosa,
  Barbu~Mititelu, Barkarson, Basile, Basmov, Batchelor, Bauer, Bedir, Behzad,
  Belieni, Bengoetxea, Benli, Ben~Moshe, Berg, Berk, Bhat, Biagetti, Bick,
  Bielinskien{\.e}, Bilgin~Ta{\c s}demir, Bjarnad{\'o}ttir, Blaschke, Blokland,
  B{\"o}bel, Bobicev, Boizou, Bonilla, Borges~V{\"o}lker, B{\"o}rstell, Bosco,
  Bouma, Bowman, Boyd, Braggaar, Branco, Brokait{\.e}, Burchardt, Cabeza,
  C{\'a}ceres~Arandia, Campos, Candito, Caron, Caron, Carvalheiro, Carvalho,
  Cassidy, Castro, Castro, Cavalcanti, Cebiro{\u g}lu~Eryi{\u g}it, Cecchini,
  Celano, {\c C}epani, {\v C}{\'e}pl{\"o}, Cesur, Cetin, {\c C}etino{\u g}lu,
  Chalub, Chamila, Chamoreau, Chauhan, Chen, Chi, Chika, Cho, Choi, Chontaeva,
  Chun, Chung, Cignarella, Cinkov{\'a}, Collomb, {\c C}{\"o}ltekin, Connor,
  Corbetta, Corbetta, Costa, Courtin, Crabb{\'e}, Cristescu, Cvetkoski, Dahan,
  Dale, Daniel, Davidson, de~Alencar, Dehouck, de~Laurentiis, de~Marneffe,
  de~Paiva, Derin, de~Souza, Diaz~de Ilarraza, D{\'{\i}}az~Hern{\'a}ndez,
  Dickerson, Di~Felippo, Dinakaramani, Di~Nuovo, Dione, Dirix, Do, Dobrovoljc,
  D{\"o}hmer, Doyle, Dozat, Droganova, Duran, Dwivedi, Ebert, Eckhoff, Eguchi,
  Eiche, Eiselen, Eli, Elkahky, Ephrem, Erina, Erjavec, Eslami, Essaidi,
  Etienne, Evelyn, Facundes, Farkas, Faryad, Favero, Ferdaousi, Fernanda,
  Fernandez~Alcalde, Fethi, Foster, Fransen, Freitas, Fujita, Gajdo{\v
  s}ov{\'a}, Galbraith, Galy, Gamba, Garcia, Garc{\'{\i}}a-Miguel,
  G{\"a}rdenfors, Gaustad, Gen{\c c}, Gerardi, Gerdes, Gessler, Ginter, Godoy,
  Goenaga, Gojenola, G{\"o}k{\i}rmak, Goldberg, Goldin, G{\'o}mez~Guinovart,
  Gonz{\'a}lez~Saavedra, Grici{\=u}t{\.e}, Grioni, Grobol, Gr{\=
  u}z{\={\i}}tis, Guillaume, Guiller, Guillot-Barbance, G{\"u}ng{\"o}r,
  Gurevich, Habash, Hafsteinsson, Haji{\v c}, Haji{\v c}~jr.,
  H{\"a}m{\"a}l{\"a}inen, H{\`a}~M{\~y}, Han, Hanifmuti, Harada, Hardwick,
  Harris, Hassert, Haug, Heinecke, Hellwig, Hennig, Hladk{\'a}, Hlav{\'a}{\v
  c}ov{\'a}, Hociung, Hoefels, Hohle, Howell, Huang, Huerta~Mendez, Hwang,
  Ikeda, Iliadou, Ingason, Ion, Irimia, Ishola, Islamaj, Ito, Iurescia,
  Jagodzi{\'n}ska, Jannat, Jel{\'{\i}}nek, Jha, Jiang, Jobanputra, Johannsen,
  J{\'o}nsd{\'o}ttir, J{\o}rgensen, Juutinen, Ka{\c s}{\i}kara, Kabaeva,
  Kahane, Kanayama, Kanerva, Kara, Karah{\'o}ǧa, K{\aa}sen, Kayadelen,
  Kengatharaiyer, Kettnerov{\'a}, Kharatyan, Kirchner, Klementieva, Klyachko,
  Kocharov, K{\"o}hn, K{\"o}ksal, Kopacewicz, Korkiakangas, K{\"o}se, Koshevoy,
  Kote, Kotsyba, Kova{\v c}i{\'c}, Kovalevskait{\.e}, Kowner, Krek,
  Krishnamurthy, K{\"u}bler, Kuqi, Kuyruk{\c c}u, Kuzgun, Kwak, Kyle, Laan,
  Laippala, Lambertino, Landau, Lando, Larasati, Lavrentiev, Lee,
  L{\^e}~H{\`{\^o}}ng, Lenci, Lertpradit, Leung, Levina, Levine, Li, Li, Li,
  Li, Li, Lim, Lima~Padovani, Lin, Lind{\'e}n, Liu, Ljube{\v s}i{\'c},
  Lobzhanidze, Loginova, Lopes, Luftiu, Lukashevskyi, Lusito, Lutgen, Luthfi,
  Luukko, Lyashevskaya, Lynn, Macketanz, Mahamdi, Maillard, Makarchuk,
  Makazhanov, Mambrini, Mandl, Manning, Manurung, Mar{\c s}an, M{\u a}r{\u
  a}nduc, Mare{\v c}ek, Marheinecke, Markantonatou, Mart{\'{\i}}nez~Alonso,
  Mart{\'{\i}}n~Rodr{\'{\i}}guez, Martins, Martins, Ma{\v s}ek, Matsuda,
  Matsumoto, Mazzei, {McDonald}, {McGuinness}, Mehta, M{\'e}nard, Mendon{\c
  c}a, Merhav, Merzhevich, Meurer, Miekka, Milano, Miller, Minerbi,
  Mischenkova, Missil{\"a}, Mititelu, Mitrofan, Miyao, Mojiri~Foroushani,
  Moln{\'a}r, Moloodi, Montemagni, More, Moreno~Romero, Moretti, Mori, Morioka,
  Moro, Mortensen, Moskalevskyi, Muischnek, Munro, Murawaki, M{\"u}{\"u}risep,
  Nainwani, Nakhl{\'e}, Navarro~Hor{\~n}iacek, Nedoluzhko, Ne{\v
  s}pore-B{\=e}rzkalne, Nevaci, Nguy{\~{\^e}}n~Th{\d i}, Nguy{\~{\^e}}n Th{\d
  i}~Minh, Nikaido, Nikolaev, Nitisaroj, Norrman, Nourian, Nunes, Nurmi, Ojala,
  Ojha, {\'O}lad{\'o}ttir, Ol{\'u}{\`o}kun, Omura, Onwuegbuzia, Ordan, Osenova,
  {\"O}stling, Ott, {\O}vrelid, {\"O}zate{\c s}, {\"O}z{\c c}elik,
  {\"O}zg{\"u}r, {\"O}zt{\"u}rk~Ba{\c s}aran, Paccosi, Palmero~Aprosio, Panova,
  Pardo, Park, Partanen, Pascual, Passarotti, Patejuk, Paulino-Passos,
  Pedonese, Peeters, Peljak-{\L}api{\'n}ska, Peng, Peng, Pereira, Pereira,
  Perez, Perkova, Perrier, Petrov, Petrova, Peverelli, Phelan, Pierre-Louis,
  Piitulainen, Pinter, Pinto, Pintucci, Pirinen, Pitler, Plamada, Plank, Plum,
  Poibeau, Ponomareva, Popel, Pretkalni{\c n}a, Pretorius, Pr{\'e}vost,
  Prokopidis, Przepi{\'o}rkowski, Pugh, Puolakainen, Purschke, Pyysalo, Qi,
  Querido, R{\"a}{\"a}bis, Rabinovich, Rademaker, Rahoman, Rama, Ramasamy,
  Ramisch, Ramos, Rashel, Rasooli, Ravishankar, Real, Rebeja, Reddy, Regnault,
  Rehm, Riabi, Riabov, Rie{\ss}ler, Rimkut{\.e}, Rinaldi, Rituma, Rizqiyah,
  Rocha, R{\"o}gnvaldsson, Roksandic, Roman, Romanenko, Rosa, Roșca, Roulon,
  Rovati, Rozonoyer, Rudina, Rueter, Ruffolo, R{\'u}narsson, Rushiti, Sadde,
  Safari, Sahala, Saleh, Salomoni, Samard{\v z}i{\'c}, Sampanis, Samson,
  S{\'a}nchez-Rodr{\'{\i}}guez, Sanguinetti, San{\i}yar, S{\"a}rg, Sartor,
  Sarymsakova, Sasaki, Saul{\={\i}}te, Savary, Sawanakunanon, Saxena, Scannell,
  Scarlata, Schang, Schneider, Schuster, Schwartz, Seddah, Seeker, Sellmer,
  Seraji, Shahzadi, Shen, Shimada, Shin, Shirasu, Shishkina, Shohibussirri,
  Shvedova, Siewert, Sigurðsson, Silva, Silveira, Silveira, Silveira, Simi,
  Simionescu, Simk{\'o}, {\v S}imkov{\'a}, S{\'{\i}}monarson, Simov,
  Sitchinava, Sither, Smith, Soares-Bastos, Solberg, Sonnenhauser, Sourov,
  Sprugnoli, Stamou, Steingr{\'{\i}}msson, Stella, Stephen, Straka, Strass,
  Strickland, Strnadov{\'a}, Suhr, Sulestio, Sulubacak, Sung, Suzuki, Swanson,
  Sz{\'a}nt{\'o}, Taguchi, Taji, Talamo, Tamburini, Tan, Tanaka, Tanaya,
  Tavoni, Tella, Tellier, Testori, Thomas, T{\i}ra{\c s}, Tonelli, Torga,
  Toska, Trosterud, Trukhina, Tsarfaty, T{\"u}rk, Tyers, {\t H}{\'o}rðarson,
  {\t H}orsteinsson, Uematsu, Untilov, Ure{\v s}ov{\'a}, Uria, Uszkoreit, Utka,
  Vagnoni, Vajjala, Vak, van~der Goot, Vanhove, van Niekerk, van Noord, Varga,
  Vedenina, Venturi, Villemonte de~la Clergerie, Vincze, Vissamsetty, Vlasova,
  Vligouridou, Wakasa, Wallenberg, Wallin, Walsh, Wang, Washington,
  Weissweiler, Wendt, Widmer, Wigderson, Wijono, Wille, Williams, Winkler,
  Wintner, Wir{\'e}n, Wittern, Woldemariam, Wong, Wr{\'o}blewska, Wu, Yako,
  Yamashita, Yamazaki, Yan, Yasuoka, Yavrumyan, Yenice, Y{\i}landilo{\u g}lu,
  Y{\i}ld{\i}z, Yu, Yuliawati, {\v Z}abokrtsk{\'y}, Zahra, Zeldes, Zhou, Zhu,
  Zhu, Zhuravleva, Ziane, and Znoti{\c n}{\v s}}]{11234/1-5787}
Daniel Zeman, Joakim Nivre, Mitchell Abrams, Elia Ackermann, No{\"e}mi Aepli,
  Hamid Aghaei, {\v Z}eljko Agi{\'c}, Amir Ahmadi, Lars Ahrenberg,
  Chika~Kennedy Ajede, Arofat Akhundjanova, Furkan Akkurt, Gabriel{\.e}
  Aleksandravi{\v c}i{\=u}t{\.e}, Ika Alfina, Avner Algom, Khalid Alnajjar,
  Chiara Alzetta, Erik Andersen, Matthew Andrews, and 633 others. 2024.
\newblock \href {http://hdl.handle.net/11234/1-5787} {Universal dependencies
  2.15}.
\newblock {LINDAT}/{CLARIAH}-{CZ} digital library at the Institute of Formal
  and Applied Linguistics ({{\'U}FAL}), Faculty of Mathematics and Physics,
  Charles University.

\bibitem[{Zhang et~al.(2020)Zhang, Li, and Zhang}]{zhang-etal-2020-efficient}
Yu~Zhang, Zhenghua Li, and Min Zhang. 2020.
\newblock \href {https://doi.org/10.18653/v1/2020.acl-main.302} {Efficient
  second-order {T}ree{CRF} for neural dependency parsing}.
\newblock In \emph{Proceedings of the 58th Annual Meeting of the Association
  for Computational Linguistics}, pages 3295--3305, Online. Association for
  Computational Linguistics.

\bibitem[{Zheng et~al.(2015)Zheng, Jayasumana, Romera-Paredes, Vineet, Su, Du,
  Huang, and Torr}]{zheng-2015-condit-random}
Shuai Zheng, Sadeep Jayasumana, Bernardino Romera-Paredes, Vibhav Vineet,
  Zhizhong Su, Dalong Du, Chang Huang, and Philip Torr. 2015.
\newblock Conditional random fields as recurrent neural networks.
\newblock In \emph{International Conference on Computer Vision (ICCV)}.

\end{thebibliography}

\appendix
\floatplacement{table}{H}
\captionsetup[table]{skip=2pt}
\renewcommand{\arraystretch}{0.92}

\section{Experimental Setup}
\label{sec:experimental-details}

\subsection{Tasks, Label Schemes, and Dataset Statistics}
\label{sec:dataset-statistics}

\paragraph{Boundary schemes.}
BMES stands for \emph{begin}, \emph{middle}, \emph{end}, and
\emph{singleton}: \texttt{S} marks a singleton object, while a multi-position
object is encoded as \texttt{B}, zero or more \texttt{M} labels, and
\texttt{E}.  NER adds the entity type to each boundary label and uses
\texttt{O} outside entities; Chinese instead adds the word's XPOS tag to every
character and has no outside label.  Thus, a valid multi-token \texttt{PER}
entity is \texttt{B-PER} (\texttt{M-PER})$^*$ \texttt{E-PER}, while a valid
multi-character noun is \texttt{B-NN} (\texttt{M-NN})$^*$ \texttt{E-NN}.
The released CoNLL-2003 BIO annotations are converted to BMES before training,
and all reported NER systems use BMES.  For Chinese, \sys{DiffusionSL} uses
BMES while our implementation uses the equivalent SBIE alphabet, whose
\texttt{I} label has exactly the role of \texttt{M}; evaluation treats
\texttt{I} and \texttt{M} identically.  POS has no boundary scheme.

Tables~\ref{tab:dataset-splits}--\ref{tab:dataset-label-distributions} are
computed from the exact preprocessed JSON files used by the experiments,
before model-specific subword tokenisation or truncation.  A unit is a word
for POS and NER and a character for Chinese Seg+POS.  The surface-form
vocabularies are case-sensitive.  Validation found no empty examples,
unit--label length mismatches, or development/test labels absent from
training.  We preserve the released splits without deduplication, so
``Dup.'' counts repeat instances beyond the first occurrence of an exact unit
sequence within each split.

\begin{table}
\centering
\scriptsize
\setlength{\tabcolsep}{2.2pt}
\begin{adjustbox}{max width=\columnwidth}
\begin{tabular}{llrrrrrrrr}
\toprule
Dataset & Split & Sent. & Units & Mean & Med. & P95 & Max & $|\mathcal{L}|$ & Dup. \\
\midrule
\multirow{3}{*}{EN-EWT}
 & Train & 12,544 & 204,579 & 16.3 & 14 & 39 & 159 & 17 & 810 \\
 & Dev   & 2,001  & 25,149  & 12.6 & 10 & 33 & 75  & 17 & 88 \\
 & Test  & 2,077  & 25,094  & 12.1 & 9  & 33 & 81  & 17 & 106 \\
\addlinespace
\multirow{3}{*}{DE-GSD}
 & Train & 13,814 & 263,791 & 19.1 & 17 & 38 & 115 & 17 & 1 \\
 & Dev   & 799    & 12,480  & 15.6 & 14 & 32 & 47  & 17 & 0 \\
 & Test  & 977    & 16,498  & 16.9 & 15 & 36 & 63  & 17 & 1 \\
\addlinespace
\multirow{3}{*}{FR-GSD}
 & Train & 14,450 & 354,652 & 24.5 & 22   & 50 & 390 & 16 & 0 \\
 & Dev   & 1,476  & 35,721  & 24.2 & 21.5 & 48 & 114 & 16 & 0 \\
 & Test  & 416    & 10,018  & 24.1 & 22   & 51 & 91  & 16 & 0 \\
\addlinespace
\multirow{3}{*}{NL-LassySmall}
 & Train & 13,817 & 239,941 & 17.4 & 16 & 39 & 124 & 16 & 297 \\
 & Dev   & 1,542  & 28,129  & 18.2 & 17 & 39 & 133 & 16 & 12 \\
 & Test  & 1,761  & 28,995  & 16.5 & 15 & 37 & 73  & 16 & 30 \\
\addlinespace
\multirow{3}{*}{CoNLL-2003}
 & Train & 14,041 & 203,621 & 14.5 & 10 & 37 & 113 & 17 & 1,350 \\
 & Dev   & 3,250  & 51,362  & 15.8 & 11 & 39 & 109 & 17 & 180 \\
 & Test  & 3,453  & 46,435  & 13.4 & 9  & 37 & 124 & 17 & 269 \\
\addlinespace
\multirow{3}{*}{ZH-GSDSimp}
 & Train & 3,997 & 156,297 & 39.1 & 34 & 79 & 182 & 104 & 0 \\
 & Dev   & 500   & 20,000  & 40.0 & 36 & 79 & 135 & 80  & 0 \\
 & Test  & 500   & 19,206  & 38.4 & 34 & 73 & 156 & 74  & 0 \\
\bottomrule
\end{tabular}
\end{adjustbox}
\caption{Split and sequence-length statistics.  Sent.\ is the number of
sentences; Units is the number of words, except characters for ZH-GSDSimp;
Mean, Med., P95, and Max describe units per sentence; $|\mathcal{L}|$ is the
number of labels observed in that split; Dup.\ is the number of exact
within-split repeat instances beyond the first occurrence.}
\label{tab:dataset-splits}
\end{table}

\begin{table}
\centering
\scriptsize
\setlength{\tabcolsep}{3pt}
\begin{adjustbox}{max width=\columnwidth}
\begin{tabular}{lllrcc}
\toprule
Dataset & Unit & Output space & $|V_{\mathrm{train}}|$ & OOV D/T (\%) & $H_{\mathrm{norm}}$ \\
\midrule
EN-EWT        & Word & UPOS (17)                    & 19,674 & 8.30/9.13   & .887 \\
DE-GSD        & Word & UPOS (17)                    & 49,498 & 10.50/11.71 & .834 \\
FR-GSD        & Word & UPOS (16)                    & 42,302 & 7.58/5.97   & .843 \\
NL-LassySmall & Word & UPOS (16)                    & 29,613 & 11.65/13.80 & .874 \\
CoNLL-2003    & Word & BMES${}\times4+\texttt{O}$ (17) & 23,623 & 8.36/12.18  & .304 \\
ZH-GSDSimp    & Char. & Boundary--XPOS (104)         & 3,450  & 0.61/0.41   & .726 \\
\bottomrule
\end{tabular}
\end{adjustbox}
\caption{Output-space and lexical statistics.  $|V_{\mathrm{train}}|$ is the
number of distinct raw training surface forms.  OOV D/T is the percentage of
development/test units absent from that raw training vocabulary.
$H_{\mathrm{norm}}=-\sum_\ell p(\ell)\log p(\ell)/\log|\mathcal{L}|$ is
training-label entropy normalised to $[0,1]$.}
\label{tab:dataset-output-spaces}
\end{table}

\begin{table}
\centering
\scriptsize
\setlength{\tabcolsep}{3pt}
\begin{adjustbox}{max width=\columnwidth}
\begin{tabular}{lp{4.1cm}crrrr}
\toprule
Dataset & Three most frequent training labels & Top 5 & $\geq10$ & 1--10 & .1--1 & $<.1$ \\
\midrule
EN-EWT & NOUN 17.0; PUNCT 11.5; VERB 11.0 & 57.4 & 3 & 11 & 3 & 0 \\
DE-GSD & NOUN 17.8; DET 14.1; PUNCT 13.1 & 67.1 & 5 & 7 & 3 & 2 \\
FR-GSD & NOUN 18.8; ADP 15.9; DET 15.3 & 68.9 & 4 & 8 & 3 & 1 \\
NL-LassySmall & NOUN 16.4; ADP 13.4; DET 11.9 & 63.8 & 5 & 9 & 1 & 1 \\
CoNLL-2003 & \texttt{O} 83.3; \texttt{S-LOC} 3.0; \texttt{B-PER} 2.1 & 92.4 & 1 & 8 & 7 & 1 \\
ZH-GSDSimp & \texttt{B-NN} 13.3; \texttt{E-NN} 13.3; \texttt{B-VV} 6.3 & 43.8 & 2 & 25 & 32 & 45 \\
\bottomrule
\end{tabular}
\end{adjustbox}
\caption{Training-label distributions (percentages).  Top 5 is the cumulative
mass of the five most frequent labels.  The final four columns count labels in
the indicated frequency bins, so together they sum to
$|\mathcal{L}|$.}
\label{tab:dataset-label-distributions}
\end{table}

\FloatBarrier
\subsection{Metrics and Output Reconstruction}
\label{sec:appendix-metrics}

POS accuracy includes punctuation.  CoNLL-2003 precision, recall, and F1 are
computed over complete entity spans.  Chinese segmentation F1 reconstructs word
spans from boundary labels while ignoring POS suffixes.  Gold-word XPOS projects
character predictions onto each gold word by majority vote and is therefore a
diagnostic, not inference with gold tokenisation.  Joint F1 requires both the
predicted word span and XPOS tag to match.  Aligned XPOS first aligns predicted
and gold words by character span, then scores XPOS on the aligned words; unlike
gold-word XPOS, it therefore reflects segmentation errors.  All systems are
evaluated after
reconstructing spans with the same evaluator.  For NER, the evaluator follows
\sys{DiffusionSL}: it retains singleton tags and complete
\texttt{B}(\texttt{M})$^*$\texttt{E} spans while ignoring malformed
fragments.  Chinese extraction analogously retains complete
\texttt{B}(\texttt{M}/\texttt{I})$^*$\texttt{E} words with a constant XPOS
component and singleton words, while ignoring incomplete fragments.

\begin{table}
\centering
\scriptsize
\setlength{\tabcolsep}{3pt}
\begin{tabular}{@{}p{.13\columnwidth}p{.39\columnwidth}p{.39\columnwidth}@{}}
\toprule
Task & Malformed sequence & Consequence \\
\midrule
NER &
\begin{tabular}[t]{@{}ccc@{}}
\texttt{John} & \texttt{Jr.} & \texttt{works}\\
\texttt{B-PER} & \texttt{M-PER} & \texttt{O}
\end{tabular} &
The \texttt{PER} span is never closed by \texttt{E-PER}; the incomplete
fragment is ignored. \\
\addlinespace
Chinese Seg+POS &
\begin{tabular}[t]{@{}ccc@{}}
$c_1$ & $c_2$ & $c_3$\\
\texttt{B-NN} & \texttt{E-VV} & \texttt{S-NN}
\end{tabular} &
The XPOS tag changes inside the first word; that fragment is ignored, while
the final singleton is retained. \\
\bottomrule
\end{tabular}
\caption{Examples of malformed predicted sequences.  POS is omitted because
every UPOS sequence is valid.}
\label{tab:malformed-examples}
\end{table}

\FloatBarrier
\subsection{Training Objectives and Hyperparameters}
\label{sec:hyperparameters}

\paragraph{Objectives.}
Let $\mathcal L_{\mathrm{post}}$ denote the sampled variational term from
\S\ref{sec:diffusion-model}---reconstruction at $t=1$ and posterior matching
otherwise---and $\mathcal L_{\mathrm{den}}$ clean-label token
cross-entropy under the model marginals, $\mathcal L_{\mathrm{seq}}$ LCRF sequence NLL,
and $\mathcal L_{\mathrm{marg}}$ token cross-entropy under exact LCRF
marginals.  The conventional \sys{Unigram} and \sys{Mean-Field} systems optimise
$\mathcal L_{\mathrm{den}}$, while the conventional \sys{LCRF} optimises
$\mathcal L_{\mathrm{seq}}$.  \sys{Diffusion-Uni} and \sys{Diffusion-MF} optimise
$\mathcal L_{\mathrm{den}}+
\lambda_{\mathrm{post}}\mathcal L_{\mathrm{post}}$; \sys{Diffusion-LCRF} optimises
$\lambda_{\mathrm{seq}}\mathcal L_{\mathrm{seq}}+
\lambda_{\mathrm{marg}}\mathcal L_{\mathrm{marg}}+
\lambda_{\mathrm{post}}\mathcal L_{\mathrm{post}}$.
Thus, $\mathcal L_{\mathrm{post}}$ is specific to diffusion, sequence NLL trains a
globally normalised LCRF, and marginal cross-entropy is an auxiliary
clean-label objective for the exact LCRF denoiser.  The token objective used by
\sys{Diffusion-Uni} and \sys{Diffusion-MF} directly supervises the marginals
consumed by the reverse update.  For \sys{Diffusion-LCRF}, sequence NLL instead
trains normalised sequence scores and marginal cross-entropy exposes their
token marginals to direct supervision; $\mathcal L_{\mathrm{post}}$ is retained only
when selected on development data.  Table~\ref{tab:training-objectives}
gives every nonzero coefficient.  All unlisted coefficients, including
additional auxiliary and structured-unary losses, are zero in the reported
configurations.

\begin{table}
\centering
\small
\setlength{\tabcolsep}{3.5pt}
\begin{tabular}{@{}lcccc@{}}
\toprule
System & $\mathcal L_{\rm den}$ & $\mathcal L_{\rm seq}$
       & $\mathcal L_{\rm marg}$ & $\mathcal L_{\rm post}$ \\
\midrule
\sys{Unigram}       & 1 & -- & -- & -- \\
\sys{Mean-Field}    & 1 & -- & -- & -- \\
\sys{LCRF}          & -- & 1 & -- & -- \\
\sys{Diffusion-Uni} & 1 & -- & -- & 1 \\
\sys{Diffusion-MF}  & 1 & -- & -- & 1 \\
\sys{Diffusion-LCRF} & -- & 1 & 1 & $\lambda_{\rm post}$ \\
\bottomrule
\end{tabular}

\begin{tabular}{@{}lcc@{}}
\toprule
Selected $\lambda_{\rm post}$ for \sys{Diffusion-LCRF}
  & Scratch & Pretrained \\
\midrule
POS             & 0   & 0 \\
NER             & 0.1 & 0 \\
Chinese Seg+POS & 0.1 & 0.1 \\
\bottomrule
\end{tabular}
\caption{Training objectives and selected \sys{Diffusion-LCRF} posterior-loss
weights.  Dashes denote absent terms; all displayed non-posterior weights are
one.}
\label{tab:training-objectives}
\end{table}

\paragraph{Architectures.}
Our implementation builds on the SuPar codebase
\citep{zhang-etal-2020-efficient}.  The pretrained encoders are
\texttt{roberta-large} \citep{DBLP:journals/corr/abs-1907-11692} for POS,
\texttt{bert-large-cased} \citep{devlin-etal-2019-bert} for NER, and
\texttt{hfl/chinese-roberta-wwm-ext}
\citep{cui-etal-2021-whole-word-masking} for Chinese.  The scratch architecture
uses the encoder described in \S\ref{sec:neural-architecture}.  We group
scratch models into \textsc{XS}, \textsc{S}, \textsc{M}, \textsc{L}, and
\textsc{XL} configurations of approximately 20, 40, 60, 80, and 650 million
parameters.  For conventional systems, additional capacity is assigned to the
encoder; for diffusion systems, the scratch encoder remains fixed and
additional capacity is assigned to the denoising Transformer.  Scoring-MLP
dropout is selected on development data.

\paragraph{Diffusion and optimisation.}
All discrete diffusion systems use $T=1000$, sample one shared
$t\sim\mathcal U\{1,\ldots,T-1\}$ per sequence during training, and follow the
logarithmic base-2 reverse schedule from $T$ to zero at inference.  Where
$\lambda_{\mathrm{post}}>0$, posterior matching uses the KL
defined in \S\ref{sec:diffusion-model}.  We optimise with Adam, clip the gradient
norm at 1, and use early stopping.  Scratch runs use a task-specific maximum
of 100--1,000 epochs; pretrained encoders are fine-tuned, with the epoch budget
selected by task.
For \sys{DiffusionSL}, we retain continuous bit diffusion and tune it
separately rather than transferring settings from our categorical models.
All reported configurations use $T=1000$ diffusion steps, linear noise, and
10-step deterministic DDIM inference with $\eta=0$ and no self-conditioning.
The selected SNR scale is 2 for POS and 0.1 for NER and Chinese.  The task- and
encoder-specific search spaces and selected settings are given in
Tables~\ref{tab:diffusionsl-search} and~\ref{tab:diffusionsl-selected}.
The scratch POS \sys{Diffusion-LCRF} checkpoints use identity potential
functions and are evaluated with exact forward--backward marginals during
reverse diffusion and final Viterbi decoding.

\paragraph{Hyperparameter search.}
Searches are staged rather than full Cartesian products: we first select the
encoder and capacity, then optimisation and denoiser size, followed by
transition parameterisation, Mean-Field iterations or LCRF loss weights, and
the final decoder.  Table~\ref{tab:hyperparameter-search} gives the union of
candidate values explored by the discrete systems; each task uses the relevant
subset.  Table~\ref{tab:selected-denoiser-settings} records the selected
structural denoiser settings, while Tables~\ref{tab:scaling_testing}
and~\ref{tab:layers_testing} report the corresponding controlled POS sweeps.
For \sys{DiffusionSL}, we first tune optimisation for each task and encoder,
then compare the applicable native and structured decoding variants.  The NER
scratch run reuses the optimisation setting selected during the encoder
comparison; the other regimes use the searches summarised below.

\begin{table}
\centering
\small
\setlength{\tabcolsep}{4pt}
\begin{tabular}{@{}p{0.31\columnwidth}p{0.63\columnwidth}@{}}
\toprule
Dimension & Candidate values explored \\
\midrule
Scratch capacity & \textsc{XS}, \textsc{S}, \textsc{M}, \textsc{L}, \textsc{XL} \\
Learning rate & $5\!\times\!10^{-6}$, $7\!\times\!10^{-6}$,
  $10^{-5}$, $2\!\times\!10^{-5}$, $3\!\times\!10^{-5}$,
  $5\!\times\!10^{-4}$, $2\!\times\!10^{-3}$ \\
Token budget / accumulation & 2,500, 5,000, or 10,000 / 1, 10, 20, or 40 steps \\
Denoiser layers & 1, 2, 3, 4, 6, 8, 10, or 12 \\
Attention heads & 8 or 16 \\
Mean-Field iterations & 1, 3, 5, 10, or 15 \\
Transitions & local or global \\
LCRF posterior weight & 0, 0.01, or 0.1 \\
Scoring-MLP dropout & 0, 0.1, or 0.33 \\
Pretrained pooling & first subword or mean subword \\
Final decoder & argmax, Mask-Viterbi, or Viterbi when applicable \\
\bottomrule
\end{tabular}
\caption{Union of hyperparameter values explored for the discrete systems.
Task- and encoder-specific searches use subsets of this space rather than one
full grid.}
\label{tab:hyperparameter-search}
\end{table}

\begin{table}
\centering
\small
\setlength{\tabcolsep}{3.5pt}
\begin{tabular}{@{}llcccc@{}}
\toprule
Task & Encoder & Layers & Heads & MF steps & Trans. \\
\midrule
POS & Scratch & 8/8/8 & 16/16/16 & 15 & L/L \\
    & Pretrained & 8/8/8 & 16/16/16 & 15 & L/L \\
NER & Scratch & 12/10/8 & 16/16/16 & 5 & G/G \\
    & Pretrained & 1/4/1 & 16/16/16 & 5 & G/G \\
Chinese & Scratch & 4/4/8 & 8/8/16 & 3 & G/G \\
        & Pretrained & 4/4/8 & 8/8/16 & 3 & G/G \\
\bottomrule
\end{tabular}
\caption{Selected diffusion-denoiser settings.  Layer and head counts are
listed in \sys{Diffusion-Uni}/\sys{Diffusion-MF}/\sys{Diffusion-LCRF} order;
MF steps is the number of Mean-Field updates inside \sys{Diffusion-MF}.
Transitions are listed for \sys{Diffusion-MF}/\sys{Diffusion-LCRF}; L and G
denote local and global parameterisations, and -- denotes an unreported model.}
\label{tab:selected-denoiser-settings}
\end{table}

\begin{table}
\centering
\small
\setlength{\tabcolsep}{4pt}
\begin{tabular}{@{}p{0.30\columnwidth}p{0.64\columnwidth}@{}}
\toprule
Setting & Candidate values explored \\
\midrule
POS, scratch
  & Batch $\{16,32,64\}$; LR $\{1,2,3,5,10\}{\times}10^{-5}$;
    warmup $\{500,1000,2000\}$; weight decay
    $\{1,10,50,100\}{\times}10^{-5}$; lowercasing on/off. \\
POS, pretrained
  & Batch $\{8,16,32\}$; encoder LR $\{0.5,1,2\}{\times}10^{-5}$;
    denoiser LR $\{1,3\}{\times}10^{-5}$; warmup $\{1000,2000\}$;
    weight decay $\{1,10\}{\times}10^{-5}$. \\
NER, scratch
  & Optimisation fixed during the encoder comparison; native extraction and
    Mask-Viterbi compared during decoder selection. \\
NER, pretrained
  & Batch $\{8,16,32\}$; LR $\{0.5,1,2\}{\times}10^{-5}$; warmup
    $\{500,1000\}$; weight decay $\{1,100,1000\}{\times}10^{-5}$;
    DDIM steps $\{10,20,50,100\}$. \\
Chinese, scratch
  & Batch $\{8,16,32,64\}$; LR $\{3,5,8,10\}{\times}10^{-5}$;
    warmup $\{500,1000,2000\}$; weight decay $\{0,1,10\}{\times}10^{-5}$;
    SNR scale $\{.05,.1,.2,.5\}$; depth $\{6,8\}$; dimension
    $\{512,768\}$; encoder/FFN dropout
    $\{(.15,.05),(.25,.1),(.35,.2)\}$; self-conditioning on/off;
    linear/cosine noise. \\
Chinese, pretrained
  & Batch $\{4,8,16\}$; encoder/denoiser LR
    $\{0.5,1,2\}{\times}10^{-5}$; warmup $\{250,1000,2000\}$;
    weight decay $\{1,1000\}{\times}10^{-5}$; SNR scale
    $\{.05,.1,.2\}$; depth $\{4,6,8\}$; self-conditioning on/off. \\
Chinese decoding
  & DDIM steps $\{10,20,50\}$; $\eta\in\{0,.5,1\}$; one or five
    samples; bit or nearest-code projection; native extraction or
    Mask-Viterbi with temperature $\{.01,.02,.05\}$, transition scale
    $\{.5,1,2\}$, and boundary scores on/off. \\
\bottomrule
\end{tabular}
\caption{\sys{DiffusionSL} search spaces.  Searches are staged and use the
listed task-relevant subsets, not their Cartesian product.  LR denotes both
encoder and denoiser learning rates unless they are listed separately.}
\label{tab:diffusionsl-search}
\end{table}

\begin{table}
\centering
\small
\setlength{\tabcolsep}{2pt}
\begin{tabular}{@{}llp{0.46\columnwidth}c@{}}
\toprule
Task & Encoder & Selected optimisation & $d$/layers \\
\midrule
POS & Scratch$^{a}$ & $b=32/64$; (LR$_e$,LR$_d)=(5,5){\times}10^{-5}$;
  warmup 1000; WD $10^{-4}$ & 512/6 \\
    & Pretrained$^{b}$ & $b=32/16$; (LR$_e$,LR$_d)=(1,3){\times}10^{-5}$;
  warmup 1000/2000; WD $10^{-4}$ & 1024/6 \\
NER & Scratch & $b=32$; (LR$_e$,LR$_d)=(5,5){\times}10^{-5}$;
  warmup 1000; WD $10^{-4}$ & 512/6 \\
    & Pretrained & $b=16$; (LR$_e$,LR$_d)=(1,1){\times}10^{-5}$;
  warmup 1000; WD $10^{-5}$ & 1024/6 \\
Chinese & Scratch & $b=8$; (LR$_e$,LR$_d)=(1,1){\times}10^{-4}$;
  warmup 1000; WD $10^{-4}$ & 512/6 \\
        & Pretrained & $b=8$; (LR$_e$,LR$_d)=(1,1){\times}10^{-5}$;
  warmup 1000; WD $10^{-5}$ & 768/6 \\
\bottomrule
\end{tabular}
\caption{Selected \sys{DiffusionSL} settings used for the reported results.
LR$_e$ and LR$_d$ are encoder and denoiser learning rates, $b$ is batch size, WD is
weight decay, and $d$ is the denoiser dimension.  All select native decoding.  $^{a}$English,
German, and Dutch use batch 32 and lowercasing; French uses batch 64 without
lowercasing.  $^{b}$English, German, and French use batch 32 and 1,000 warmup
steps; Dutch uses batch 16 and 2,000 steps.}
\label{tab:diffusionsl-selected}
\end{table}

\FloatBarrier
\subsection{Training, Selection, and Aggregation}
\label{sec:selection-aggregation}

Each candidate configuration is compared over four random seeds using POS
accuracy, NER entity F1, or Chinese joint F1 on the development set.  For each
system and encoder, we select the hyperparameters and decoding rule by this
mean development score, then report the corresponding four-seed test result.
The secondary Chinese metrics come from the same selected predictions.  All
reported deviations are sample standard deviations across the four seeds.

\FloatBarrier
\subsection{Software, Compute, and Artefact Release}
\label{sec:software-artifacts}

\textbf{Software.} Experiments use PyTorch ($\geq2.2$), Hugging Face Transformers
($\geq4.38$), and SuPar \citep{zhang-etal-2020-efficient}.  Appendix
\ref{sec:hyperparameters} gives pretrained model identifiers and model and
optimisation settings; native tokenisers use batch padding and 256-unit
truncation.  Appendices~\ref{sec:dataset-statistics}--\ref{sec:appendix-metrics}
specify preprocessing and evaluation.  The anonymised artefact is available at
\url{https://anonymous.4open.science/r/diffusion-mf-artifact-1058/}; it includes
the environment, configurations, and evaluation scripts.
\textbf{Compute.} Training and tuning used one NVIDIA V100, A100, or H200 GPU per run.
Appendix~\ref{sec:hyperparameters} gives model-size ranges and
Appendix~\ref{sec:timing-experiments} the A100 benchmark setup.  Incomplete
historical records preclude a reliable total GPU-hour estimate.
\textbf{Licences and release.} The DiffusionSL code has no licence, so neither
it nor our Mask-Viterbi extension can be released.  The artefact releases our
other code and configurations under GPL-3.0; incorporated SuPar components
retain their MIT licence and attribution.  It will contain no datasets or
pretrained checkpoints.  Users obtain upstream resources under their terms;
conversion scripts require user-provided copies.

\FloatBarrier
\section{Inference and Decoding}
\label{sec:inference-decoding}

\subsection{Reverse-Diffusion Inference}
\label{sec:decoding-algorithms}

Algorithm~\ref{inference_algorithm} gives the unbatched reverse-diffusion
procedure with timesteps halved at each round.  The implementation propagates
a categorical distribution $\boldsymbol\pi_t$ at every position rather than a
sampled label sequence.  Its expected tag embeddings condition the denoiser,
which returns estimated clean-label marginals $\boldsymbol\rho_0$ from the
unigram, Mean-Field, or exact LCRF model.  The closed-form reverse posterior
combines $\boldsymbol\pi_t$, $\boldsymbol\rho_0$, and the forward transition
matrices to obtain the normalised distribution at
$t'=\lfloor t/2\rfloor$.  The final categorical scores are passed to the
decoder selected on development data.

\SetKwInput{KwInput}{Input}
\SetKwInput{KwOutput}{Output}
\begin{algorithm}
\DontPrintSemicolon

\KwIn{$\bm{s}=s_1\dots s_N$: Input sequence}
\KwOut{$\widehat{\bm y} \in \mathcal{L}^{N}$: Predicted tags}

$t \leftarrow T$\

$\boldsymbol\pi_t \leftarrow
  \frac{1}{|\mathcal L|}\bm{1}_{N\times |\mathcal L|}$\

\While{$t > 0$}{
    $\widetilde{y}_i \leftarrow
      \sum_{\ell\in\mathcal L}\pi_{t,i}(\ell)\,
      \text{tag-embed}(\ell)$\

    $\widetilde{t} \leftarrow \text{time-embed}(t)$\

    $\boldsymbol\rho_0\leftarrow
      \text{denoise-marginals}(\bm{s}, \widetilde{y}, \widetilde{t})$\

    $t' \leftarrow \left\lfloor \frac{t}{2} \right\rfloor$\

    $\boldsymbol\pi_{t'} \leftarrow
      \text{reverse-posterior}(\boldsymbol\pi_t,
      \boldsymbol\rho_0,t,t')$\

    $t \leftarrow t'$\
}

$\hat{y} \leftarrow \text{final-decode}(\boldsymbol\pi_0)$\\
\Return $\hat{y}$\

\caption{Reverse-diffusion inference for sequence labelling.}
\label{inference_algorithm}
\end{algorithm}

\FloatBarrier
\subsection{Final Decoders and Constraints}
\label{sec:final-decoders}

\begin{table}
  \centering
  \scriptsize
  \setlength{\tabcolsep}{3pt}
  \begin{tabular}{@{}lccc@{}}
    \toprule
    System & POS & NER & Chinese \\
    \midrule
    \sys{Unigram}       & Arg.    & Arg./M-Vit.    & Arg./M-Vit. \\
    \sys{Mean-Field}    & Arg.    & Arg./M-Vit.    & Arg./M-Vit. \\
    \sys{LCRF}          & Vit.    & Vit.           & Vit. \\
    \sys{Diffusion-Uni} & Arg.    & Arg./M-Vit.    & Arg./M-Vit. \\
    \sys{Diffusion-MF}  & Arg.    & Arg./M-Vit.    & Arg./M-Vit. \\
    \sys{Diffusion-LCRF} & Vit.   & Vit.           & Vit. \\
    \sys{DiffusionSL}   & Native  & Native/M-Vit.$^*$ & Native/M-Vit.$^*$ \\
    \bottomrule
  \end{tabular}
  \caption{Available final decoders.  Arg., M-Vit., and Vit. denote
  position-independent argmax, Mask-Viterbi with fixed legality constraints,
  and Viterbi with learned transition scores (and task constraints for NER and
  Chinese).  Native is \sys{DiffusionSL}'s threshold-and-decode rule; $^*$
  marks our distance-based unary extension.
  Mask-Viterbi is unnecessary for POS because every label sequence is valid.}
  \label{tab:decoder-compatibility}
\end{table}

Mask-Viterbi uses fixed legality scores---zero for legal transitions and
$-\infty$ for illegal ones---whereas LCRF Viterbi additionally uses learned
transition scores.  Both enforce validity; position-independent argmax may
not.  Mask-Viterbi is used only for NER and Chinese Seg+POS because every POS
label sequence is valid.

\FloatBarrier
\subsection{Mask-Viterbi Extension for
\texorpdfstring{\sys{DiffusionSL}}{DiffusionSL}}
\label{sec:diffusionsl-mask-viterbi}

\paragraph{Native decoding.}
Let $K=|\mathcal L|$ and $b=\lceil\log_2 K\rceil$.  \sys{DiffusionSL}
assigns each label index $\ell\in\{0,\ldots,K-1\}$ a scaled binary code
\[
  \mathbf c_\ell
  =s\bigl(2\,\operatorname{bin}_b(\ell)-\mathbf 1\bigr)
  \in\{-s,+s\}^{b},
\]
where $s$ is the scale used during diffusion.  After reverse diffusion, the
model returns a continuous vector $\mathbf z_i\in\mathbb R^b$ at each
position.  Its native decoder thresholds every coordinate and converts the
resulting bits back to an integer:
\[
  \widehat y_i^{\mathrm{native}}
  =\operatorname{dec}_b\!\left(\mathbb I[\mathbf z_i>0]\right).
\]
This rule produces one label independently at each position, but does not
provide scores or a ranking over alternative labels.  The \emph{Arg.} rows
for \sys{DiffusionSL} in the full-result tables refer to this native
positionwise decoder rather than an argmax over categorical logits.

\paragraph{From bit vectors to unary scores.}
We leave the trained checkpoint and the entire reverse-diffusion trajectory
unchanged.  Only after the final denoising step do we compare
$\mathbf z_i$ with every valid label code and define
\[
  u_i(\ell)
  =-\frac{\lVert\mathbf z_i-\mathbf c_\ell\rVert_2^2}{\tau},
  \qquad \ell\in\mathcal L ,
\]
with temperature $\tau=0.05$.  A softmax is unnecessary because Viterbi
depends only on relative scores.  If all $2^b$ bit patterns correspond to
labels, independent maximisation of $u_i$ is equivalent to coordinate-wise
thresholding.  When $K<2^b$, it instead chooses the nearest code among the
$K$ valid labels if the thresholded pattern is unused.  In either case, the
construction supplies the complete unary table required by structured
decoding; it should be viewed as a distance-based surrogate, not as a
categorical distribution learned by \sys{DiffusionSL}.

\paragraph{Fixed constraints.}
For NER and Chinese Seg+POS, let $A(\ell,\ell')$ be zero when the BMES
transition $\ell\!\rightarrow\!\ell'$ is legal and $-\infty$ otherwise.
Analogous start and end scores $a_{\mathrm s}$ and $a_{\mathrm e}$ exclude
invalid boundary tags.  Mask-Viterbi returns
\[
\begin{aligned}
  \widehat{\mathbf y}
  =\argmax_{\mathbf y\in\mathcal L^n}\biggl\{&
  a_{\mathrm s}(y_1)+\sum_{i=1}^{n}u_i(y_i)\\
  &+\sum_{i=2}^{n}A(y_{i-1},y_i)+a_{\mathrm e}(y_n)
  \biggr\}.
\end{aligned}
\]
NER spans must begin and end legally, and \texttt{M}/\texttt{E} tags must
retain the entity type opened by the preceding \texttt{B}/\texttt{M} tag.
Chinese uses the same boundary rules, with the XPOS component held constant
within a word.  POS has no corresponding hard grammar, so this decoder is
not used for POS.  Because all legal transitions receive the same score,
Mask-Viterbi enforces validity without introducing learned transition
preferences.

\paragraph{Implementation and effect.}
The extension is inference-only and adds no parameters.  Constructing the
unaries costs $O(nKb)$ and Viterbi costs $O(nK^2)$ time with $O(nK)$
backpointers.  We use fixed transition, start, and end masks, representing
$-\infty$ numerically by $-10^4$.  The constrained and native rows therefore
use the same trained models and differ only at final decoding.

The extension guarantees a valid BMES sequence, but lowers
\sys{DiffusionSL}'s primary metric for both encoders on both constrained
tasks (Tables~\ref{tab:ner-scratch-full}, \ref{tab:ner-bert-full},
\ref{tab:segpos-scratch-full}, and \ref{tab:segpos-pretrained-full}).
This can occur because Euclidean distance in bit space is not trained as a
calibrated score over labels, while the hard projection may replace a locally
preferred tag to repair the sequence.  Our discrete diffusion models instead
produce categorical scores directly, so the same fixed constraints can be
applied without first introducing this post-hoc scoring surrogate.

\FloatBarrier
\section{Complete Results}
\label{sec:full-results}

\subsection{Multilingual POS Tagging}
\label{sec:full-pos-results}
\begin{table}
\centering
\scriptsize
\setlength{\tabcolsep}{3.2pt}
\begin{tabular}{lccccc}
\toprule
Model & EN & DE & FR & NL & Avg. \\
\midrule
\multicolumn{6}{l}{\emph{Scratch encoder}} \\
\sys{Unigram} & 91.48 & 92.38 & 94.58 & 91.94 & 92.60 \\
\sys{Mean-Field} & 93.51 & 93.88 & 96.02 & 93.84 & 94.31 \\
\sys{LCRF} & 94.02 & 94.11 & 96.60 & 94.00 & 94.68 \\
\sys{DiffusionSL} & 93.83 & 94.29 & 96.67 & 93.49 & 94.57 \\
\sys{Diffusion-Uni} & 95.01 & 94.73 & 97.33 & 94.73 & 95.45 \\
\sys{Diffusion-LCRF} & 91.30 & 93.86 & 93.72 & 90.48 & 92.34 \\
\sys{Diffusion-MF} & \textbf{95.06} & \textbf{94.85} & \textbf{97.39} & \textbf{94.93} & \textbf{95.56} \\
\midrule
\multicolumn{6}{l}{\emph{RoBERTa-large}} \\
\sys{Unigram} & 98.41 & 96.91 & 98.26 & 97.56 & 97.78 \\
\sys{Mean-Field} & 98.28 & 96.80 & 98.14 & 97.35 & 97.64 \\
\sys{LCRF} & 98.41 & 96.85 & 98.26 & 97.48 & 97.75 \\
\sys{DiffusionSL} & 98.30 & 96.77 & 98.26 & 97.44 & 97.69 \\
\sys{Diffusion-Uni} & \textbf{98.51} & \textbf{96.90} & 98.27 & 97.59 & 97.82 \\
\sys{Diffusion-LCRF} & 98.38 & 96.89 & \textbf{98.33} & 97.50 & 97.77 \\
\sys{Diffusion-MF} & 98.50 & 96.89 & 98.30 & \textbf{97.62} & \textbf{97.83} \\
\bottomrule
\end{tabular}
\caption{POS test accuracy on UD~2.15.  Avg.\ is the unweighted mean of
the four language means.  All entries are means over four seeds.}
\label{tab:full-pos}
\end{table}

\FloatBarrier
\subsection{CoNLL-2003 Named-Entity Recognition}
\label{sec:ner-results}
\begin{table}
\centering
\scriptsize
\setlength{\tabcolsep}{1.8pt}
\begin{adjustbox}{max width=\columnwidth}
\begin{tabular}{llcccc}
\toprule
Model & Dec. & Acc. & P & R & F1 \\
\midrule
\sys{DiffusionSL} & Arg.$^\dagger$ & $97.12{\pm}.04$ & $88.49{\pm}.41$ & $83.57{\pm}.26$ & $85.96{\pm}.23$ \\
\sys{DiffusionSL} & M-Vit. & $97.16{\pm}.02$ & $86.43{\pm}.37$ & $84.62{\pm}.09$ & $85.51{\pm}.21$ \\
\midrule
\sys{Unigram} & Arg. & $95.53{\pm}.06$ & $87.65{\pm}.58$ & $70.21{\pm}.61$ & $77.96{\pm}.40$ \\
\sys{Unigram} & M-Vit.$^\dagger$ & $96.34{\pm}.07$ & $82.04{\pm}1.21$ & $79.56{\pm}.90$ & $80.76{\pm}.18$ \\
\sys{Mean-Field} & Arg.$^\dagger$ & $96.84{\pm}.05$ & $84.89{\pm}.78$ & $81.45{\pm}.45$ & $83.13{\pm}.40$ \\
\sys{Mean-Field} & M-Vit. & $96.84{\pm}.05$ & $83.26{\pm}.50$ & $82.56{\pm}.12$ & $82.91{\pm}.19$ \\
\sys{LCRF} & Vit.$^\dagger$ & $97.31{\pm}.05$ & $87.31{\pm}.06$ & $84.51{\pm}.40$ & $85.89{\pm}.19$ \\
\midrule
\sys{Diffusion-Uni} & Arg.$^\dagger$ & $97.47{\pm}.10$ & $88.81{\pm}.29$ & $84.85{\pm}.83$ & $86.78{\pm}.53$ \\
\sys{Diffusion-Uni} & M-Vit. & $97.48{\pm}.04$ & $87.73{\pm}.42$ & $85.61{\pm}.17$ & $86.66{\pm}.26$ \\
\sys{Diffusion-MF} & Arg. & $97.43{\pm}.16$ & $88.84{\pm}.48$ & $84.66{\pm}1.28$ & $86.70{\pm}.83$ \\
\sys{Diffusion-MF} & M-Vit.$^\dagger$ & $97.54{\pm}.13$ & $87.95{\pm}.32$ & $86.01{\pm}.96$ & $\mathbf{86.97{\pm}.60}$ \\
\sys{Diffusion-LCRF} & Vit.$^\dagger$ & $97.47{\pm}.07$ & $87.99{\pm}.79$ & $85.41{\pm}.25$ & $86.68{\pm}.45$ \\
\bottomrule
\end{tabular}
\end{adjustbox}
\caption{CoNLL-2003 development results with the scratch encoder.  Values
are mean $\pm$ standard deviation over four seeds.  Arg., M-Vit., and Vit.\
denote argmax, mask-Viterbi, and learned-transition Viterbi; $^\dagger$ marks
the development-selected decoder.}
\label{tab:ner-scratch-dev-full}
\end{table}

\begin{table}
\centering
\scriptsize
\setlength{\tabcolsep}{1.8pt}
\begin{adjustbox}{max width=\columnwidth}
\begin{tabular}{llcccc}
\toprule
Model & Dec. & Acc. & P & R & F1 \\
\midrule
\sys{DiffusionSL} & Arg.$^\dagger$ & $95.70{\pm}.09$ & $82.34{\pm}.48$ & $77.48{\pm}.61$ & $79.84{\pm}.50$ \\
\sys{DiffusionSL} & M-Vit. & $95.73{\pm}.12$ & $80.27{\pm}.71$ & $78.52{\pm}.58$ & $79.39{\pm}.59$ \\
\midrule
\sys{Unigram} & Arg. & $93.66{\pm}.07$ & $79.77{\pm}1.01$ & $61.93{\pm}.88$ & $69.72{\pm}.54$ \\
\sys{Unigram} & M-Vit.$^\dagger$ & $94.84{\pm}.07$ & $74.89{\pm}1.87$ & $72.70{\pm}1.32$ & $73.75{\pm}.49$ \\
\sys{Mean-Field} & Arg.$^\dagger$ & $95.38{\pm}.10$ & $77.73{\pm}.85$ & $74.93{\pm}.34$ & $76.31{\pm}.52$ \\
\sys{Mean-Field} & M-Vit. & $95.41{\pm}.09$ & $76.40{\pm}.50$ & $76.37{\pm}.60$ & $76.39{\pm}.46$ \\
\sys{LCRF} & Vit.$^\dagger$ & $95.73{\pm}.10$ & $80.02{\pm}.45$ & $77.80{\pm}.44$ & $78.89{\pm}.42$ \\
\midrule
\sys{Diffusion-Uni} & Arg.$^\dagger$ & $95.98{\pm}.18$ & $82.43{\pm}.62$ & $77.97{\pm}1.26$ & $80.14{\pm}.86$ \\
\sys{Diffusion-Uni} & M-Vit. & $96.05{\pm}.04$ & $81.37{\pm}.65$ & $79.28{\pm}.39$ & $80.31{\pm}.20$ \\
\sys{Diffusion-MF} & Arg. & $96.04{\pm}.17$ & $82.71{\pm}.50$ & $78.43{\pm}1.22$ & $80.51{\pm}.76$ \\
\sys{Diffusion-MF} & M-Vit.$^\dagger$ & $96.15{\pm}.16$ & $81.46{\pm}.44$ & $79.86{\pm}1.08$ & $\mathbf{80.65{\pm}.69}$ \\
\sys{Diffusion-LCRF} & Vit.$^\dagger$ & $95.99{\pm}.08$ & $81.33{\pm}1.29$ & $78.38{\pm}.57$ & $79.83{\pm}.51$ \\
\bottomrule
\end{tabular}
\end{adjustbox}
\caption{CoNLL-2003 test results with the scratch encoder.  Values are mean
$\pm$ standard deviation over four seeds; P, R, and F1 are entity-level, and
$^\dagger$ marks the development-selected decoder.}
\label{tab:ner-scratch-full}
\end{table}

\begin{table}
\centering
\scriptsize
\setlength{\tabcolsep}{1.8pt}
\begin{adjustbox}{max width=\columnwidth}
\begin{tabular}{llcccc}
\toprule
Model & Dec. & Acc. & P & R & F1 \\
\midrule
\sys{DiffusionSL} & Arg.$^\dagger$ & $99.18{\pm}.02$ & $96.49{\pm}.13$ & $95.96{\pm}.11$ & $96.23{\pm}.08$ \\
\sys{DiffusionSL} & M-Vit. & $99.17{\pm}.02$ & $95.97{\pm}.13$ & $96.06{\pm}.08$ & $96.02{\pm}.09$ \\
\midrule
\sys{Unigram} & Arg. & $99.18{\pm}.03$ & $95.94{\pm}.13$ & $95.86{\pm}.18$ & $95.90{\pm}.10$ \\
\sys{Unigram} & M-Vit.$^\dagger$ & $99.21{\pm}.02$ & $95.79{\pm}.10$ & $96.11{\pm}.12$ & $95.95{\pm}.05$ \\
\sys{Mean-Field} & Arg. & $99.24{\pm}.02$ & $96.20{\pm}.11$ & $96.23{\pm}.10$ & $96.22{\pm}.09$ \\
\sys{Mean-Field} & M-Vit.$^\dagger$ & $99.26{\pm}.02$ & $96.15{\pm}.11$ & $96.31{\pm}.11$ & $96.23{\pm}.10$ \\
\sys{LCRF} & Vit.$^\dagger$ & $99.23{\pm}.01$ & $96.09{\pm}.04$ & $96.22{\pm}.06$ & $96.16{\pm}.04$ \\
\midrule
\sys{Diffusion-Uni} & Arg.$^\dagger$ & $99.20{\pm}.04$ & $96.29{\pm}.03$ & $96.09{\pm}.14$ & $96.19{\pm}.07$ \\
\sys{Diffusion-Uni} & M-Vit. & $99.25{\pm}.02$ & $96.06{\pm}.05$ & $96.25{\pm}.15$ & $96.15{\pm}.09$ \\
\sys{Diffusion-MF} & Arg.$^\dagger$ & $99.21{\pm}.03$ & $96.34{\pm}.07$ & $96.22{\pm}.12$ & $96.28{\pm}.07$ \\
\sys{Diffusion-MF} & M-Vit. & $99.27{\pm}.02$ & $96.12{\pm}.10$ & $96.42{\pm}.17$ & $96.27{\pm}.12$ \\
\sys{Diffusion-LCRF} & Vit.$^\dagger$ & $99.30{\pm}.01$ & $96.29{\pm}.09$ & $96.50{\pm}.02$ & $\mathbf{96.39{\pm}.05}$ \\
\bottomrule
\end{tabular}
\end{adjustbox}
\caption{CoNLL-2003 development results with \texttt{bert-large-cased}.
Values are mean $\pm$ standard deviation over four seeds; $^\dagger$ marks
the development-selected decoder.}
\label{tab:ner-bert-dev-full}
\end{table}

\begin{table}
\centering
\scriptsize
\setlength{\tabcolsep}{1.8pt}
\begin{adjustbox}{max width=\columnwidth}
\begin{tabular}{llcccc}
\toprule
Model & Dec. & Acc. & P & R & F1 \\
\midrule
\sys{DiffusionSL} & Arg.$^\dagger$ & $98.29{\pm}.05$ & $92.58{\pm}.22$ & $92.18{\pm}.19$ & $92.38{\pm}.17$ \\
\sys{DiffusionSL} & M-Vit. & $98.27{\pm}.05$ & $91.87{\pm}.34$ & $92.27{\pm}.22$ & $92.07{\pm}.25$ \\
\midrule
\sys{Unigram} & Arg. & $98.28{\pm}.02$ & $91.60{\pm}.23$ & $92.00{\pm}.23$ & $91.80{\pm}.04$ \\
\sys{Unigram} & M-Vit.$^\dagger$ & $98.30{\pm}.02$ & $91.24{\pm}.13$ & $92.25{\pm}.17$ & $91.74{\pm}.05$ \\
\sys{Mean-Field} & Arg. & $98.37{\pm}.02$ & $91.85{\pm}.13$ & $92.35{\pm}.19$ & $92.10{\pm}.09$ \\
\sys{Mean-Field} & M-Vit.$^\dagger$ & $98.38{\pm}.01$ & $91.74{\pm}.10$ & $92.45{\pm}.14$ & $92.09{\pm}.04$ \\
\sys{LCRF} & Vit.$^\dagger$ & $98.38{\pm}.03$ & $91.71{\pm}.24$ & $92.50{\pm}.11$ & $92.11{\pm}.15$ \\
\midrule
\sys{Diffusion-Uni} & Arg.$^\dagger$ & $98.31{\pm}.04$ & $92.13{\pm}.11$ & $92.84{\pm}.19$ & $92.49{\pm}.11$ \\
\sys{Diffusion-Uni} & M-Vit. & $98.42{\pm}.03$ & $91.83{\pm}.07$ & $92.93{\pm}.21$ & $92.38{\pm}.12$ \\
\sys{Diffusion-MF} & Arg.$^\dagger$ & $98.30{\pm}.07$ & $92.27{\pm}.14$ & $92.75{\pm}.15$ & $\mathbf{92.51{\pm}.14}$ \\
\sys{Diffusion-MF} & M-Vit. & $98.44{\pm}.03$ & $91.99{\pm}.13$ & $92.90{\pm}.12$ & $92.44{\pm}.12$ \\
\sys{Diffusion-LCRF} & Vit.$^\dagger$ & $98.46{\pm}.05$ & $92.11{\pm}.25$ & $92.76{\pm}.10$ & $92.43{\pm}.17$ \\
\bottomrule
\end{tabular}
\end{adjustbox}
\caption{CoNLL-2003 test results with \texttt{bert-large-cased}.
Values are mean $\pm$ standard deviation over four seeds; $^\dagger$ marks
the development-selected decoder.}
\label{tab:ner-bert-full}
\end{table}

\FloatBarrier
\subsection{Chinese Segmentation and POS Tagging}
\label{sec:segpos-results}
\begin{table}
\centering
\scriptsize
\setlength{\tabcolsep}{3.5pt}
\begin{adjustbox}{max width=\columnwidth}
\begin{tabular}{lccc}
\toprule
Model & Seg.\ F1 & Gold XPOS & Joint F1 \\
\midrule
\multicolumn{4}{l}{\emph{Argmax}} \\
\sys{DiffusionSL}$^\dagger$ & $90.06{\pm}.46$ & $86.36{\pm}.33$ & $83.34{\pm}.38$ \\
\sys{Unigram} & $72.62{\pm}.27$ & $77.50{\pm}.34$ & $65.91{\pm}.42$ \\
\sys{Mean-Field} & $89.79{\pm}.33$ & $86.25{\pm}.12$ & $82.24{\pm}.19$ \\
\sys{Diffusion-Uni}$^\dagger$ & $92.73{\pm}.13$ & $89.18{\pm}.30$ & $86.39{\pm}.16$ \\
\sys{Diffusion-MF}$^\dagger$ & $92.62{\pm}.32$ & $89.22{\pm}.18$ & $\mathbf{86.46{\pm}.25}$ \\
\midrule
\multicolumn{4}{l}{\emph{Mask-Viterbi}} \\
\sys{DiffusionSL} & $89.05{\pm}.81$ & $85.68{\pm}.46$ & $81.44{\pm}.64$ \\
\sys{Unigram}$^\dagger$ & $86.18{\pm}.32$ & $82.57{\pm}.24$ & $76.39{\pm}.30$ \\
\sys{Mean-Field}$^\dagger$ & $90.49{\pm}.22$ & $86.32{\pm}.14$ & $82.46{\pm}.08$ \\
\sys{Diffusion-Uni} & $\mathbf{92.90{\pm}.16}$ & $89.24{\pm}.29$ & $86.28{\pm}.22$ \\
\sys{Diffusion-MF} & $92.83{\pm}.21$ & $\mathbf{89.34{\pm}.21}$ & $86.43{\pm}.26$ \\
\midrule
\multicolumn{4}{l}{\emph{Learned-transition Viterbi}} \\
\sys{LCRF}$^\dagger$ & $90.41{\pm}.16$ & $86.29{\pm}.35$ & $82.81{\pm}.23$ \\
\sys{Diffusion-LCRF}$^\dagger$ & $91.33{\pm}.34$ & $87.23{\pm}.24$ & $83.74{\pm}.32$ \\
\bottomrule
\end{tabular}
\end{adjustbox}
\caption{Chinese GSDSimp development results without a pretrained encoder.
Values are mean $\pm$ standard deviation over four seeds; $^\dagger$ marks
the development-selected decoder.}
\label{tab:segpos-scratch-dev-full}
\end{table}

\begin{table}
\centering
\scriptsize
\setlength{\tabcolsep}{3.5pt}
\begin{adjustbox}{max width=\columnwidth}
\begin{tabular}{lccc}
\toprule
Model & Seg.\ F1 & Gold XPOS & Joint F1 \\
\midrule
\multicolumn{4}{l}{\emph{Argmax}} \\
\sys{DiffusionSL}$^\dagger$ & $90.37{\pm}.30$ & $86.94{\pm}.27$ & $83.89{\pm}.40$ \\
\sys{Unigram} & $73.04{\pm}.65$ & $78.55{\pm}.43$ & $66.45{\pm}.67$ \\
\sys{Mean-Field} & $90.45{\pm}.18$ & $87.10{\pm}.20$ & $83.35{\pm}.32$ \\
\sys{Diffusion-Uni}$^\dagger$ & $92.76{\pm}.34$ & $89.36{\pm}.27$ & $86.52{\pm}.36$ \\
\sys{Diffusion-MF}$^\dagger$ & $92.89{\pm}.23$ & $89.50{\pm}.36$ & $\mathbf{86.69{\pm}.32}$ \\
\midrule
\multicolumn{4}{l}{\emph{Mask-Viterbi}} \\
\sys{DiffusionSL} & $89.39{\pm}.61$ & $86.35{\pm}.35$ & $82.06{\pm}.69$ \\
\sys{Unigram}$^\dagger$ & $86.25{\pm}.10$ & $83.17{\pm}.21$ & $77.01{\pm}.18$ \\
\sys{Mean-Field}$^\dagger$ & $90.84{\pm}.19$ & $87.14{\pm}.29$ & $83.34{\pm}.27$ \\
\sys{Diffusion-Uni} & $92.92{\pm}.39$ & $89.35{\pm}.18$ & $86.38{\pm}.34$ \\
\sys{Diffusion-MF} & $\mathbf{93.03{\pm}.18}$ & $\mathbf{89.52{\pm}.29}$ & $86.60{\pm}.28$ \\
\midrule
\multicolumn{4}{l}{\emph{Learned-transition Viterbi}} \\
\sys{LCRF}$^\dagger$ & $91.09{\pm}.12$ & $87.17{\pm}.09$ & $83.78{\pm}.21$ \\
\sys{Diffusion-LCRF}$^\dagger$ & $91.65{\pm}.31$ & $87.57{\pm}.25$ & $84.20{\pm}.48$ \\
\bottomrule
\end{tabular}
\end{adjustbox}
\caption{Chinese GSDSimp test results without a pretrained encoder.
Values are mean $\pm$ standard deviation over four seeds; $^\dagger$ marks
the development-selected decoder.}
\label{tab:segpos-scratch-full}
\end{table}

\begin{table}
\centering
\scriptsize
\setlength{\tabcolsep}{2.5pt}
\begin{adjustbox}{max width=\columnwidth}
\begin{tabular}{lcccc}
\toprule
Model & Seg.\ F1 & Gold XPOS & Aligned XPOS & Joint F1 \\
\midrule
\multicolumn{5}{l}{\emph{Argmax}} \\
\sys{DiffusionSL}$^\dagger$ & $97.94{\pm}.05$ & $96.37{\pm}.07$ & $97.28{\pm}.12$ & $95.40{\pm}.05$ \\
\sys{Unigram} & $98.03{\pm}.04$ & $96.26{\pm}.05$ & $96.91{\pm}.05$ & $94.99{\pm}.10$ \\
\sys{Mean-Field} & $97.95{\pm}.05$ & $96.28{\pm}.04$ & $97.15{\pm}.04$ & $95.04{\pm}.06$ \\
\sys{Diffusion-Uni} & $98.03{\pm}.08$ & $96.53{\pm}.03$ & $97.29{\pm}.07$ & $95.30{\pm}.05$ \\
\sys{Diffusion-MF} & $98.02{\pm}.05$ & $96.53{\pm}.03$ & $97.33{\pm}.02$ & $95.34{\pm}.06$ \\
\midrule
\multicolumn{5}{l}{\emph{Mask-Viterbi}} \\
\sys{DiffusionSL} & $97.87{\pm}.05$ & $96.26{\pm}.06$ & $97.16{\pm}.03$ & $95.09{\pm}.06$ \\
\sys{Unigram}$^\dagger$ & $98.09{\pm}.01$ & $96.31{\pm}.04$ & $97.03{\pm}.03$ & $95.18{\pm}.03$ \\
\sys{Mean-Field}$^\dagger$ & $98.10{\pm}.02$ & $96.32{\pm}.04$ & $97.05{\pm}.03$ & $95.20{\pm}.05$ \\
\sys{Diffusion-Uni}$^\dagger$ & $98.02{\pm}.08$ & $96.53{\pm}.04$ & $\mathbf{97.36{\pm}.07}$ & $95.43{\pm}.06$ \\
\sys{Diffusion-MF}$^\dagger$ & $98.05{\pm}.05$ & $\mathbf{96.54{\pm}.04}$ & $97.35{\pm}.03$ & $\mathbf{95.44{\pm}.03}$ \\
\midrule
\multicolumn{5}{l}{\emph{Learned-transition Viterbi}} \\
\sys{LCRF}$^\dagger$ & $\mathbf{98.11{\pm}.02}$ & $96.28{\pm}.04$ & $96.99{\pm}.03$ & $95.16{\pm}.02$ \\
\sys{Diffusion-LCRF}$^\dagger$ & $97.98{\pm}.11$ & $96.36{\pm}.12$ & $97.13{\pm}.13$ & $95.17{\pm}.24$ \\
\bottomrule
\end{tabular}
\end{adjustbox}
\caption{Chinese GSDSimp development results with
\texttt{hfl/chinese-roberta-wwm-ext}.  Aligned XPOS is computed after
aligning predicted and gold word spans.  Values are mean $\pm$ standard
deviation over four seeds; $^\dagger$ marks the development-selected decoder.}
\label{tab:segpos-pretrained-dev-full}
\end{table}

\begin{table}
\centering
\scriptsize
\setlength{\tabcolsep}{2.5pt}
\begin{adjustbox}{max width=\columnwidth}
\begin{tabular}{lcccc}
\toprule
Model & Seg.\ F1 & Gold XPOS & Aligned XPOS & Joint F1 \\
\midrule
\multicolumn{5}{l}{\emph{Argmax}} \\
\sys{DiffusionSL}$^\dagger$ & $98.09{\pm}.07$ & $96.62{\pm}.07$ & $97.36{\pm}.10$ & $95.59{\pm}.06$ \\
\sys{Unigram} & $97.99{\pm}.05$ & $96.42{\pm}.04$ & $97.06{\pm}.01$ & $95.03{\pm}.08$ \\
\sys{Mean-Field} & $97.92{\pm}.09$ & $96.40{\pm}.06$ & $97.28{\pm}.05$ & $95.06{\pm}.09$ \\
\sys{Diffusion-Uni} & $98.21{\pm}.09$ & $96.72{\pm}.04$ & $97.35{\pm}.04$ & $95.52{\pm}.12$ \\
\sys{Diffusion-MF} & $98.21{\pm}.04$ & $96.73{\pm}.06$ & $97.37{\pm}.06$ & $95.57{\pm}.08$ \\
\midrule
\multicolumn{5}{l}{\emph{Mask-Viterbi}} \\
\sys{DiffusionSL} & $98.09{\pm}.04$ & $96.58{\pm}.07$ & $97.27{\pm}.08$ & $95.41{\pm}.07$ \\
\sys{Unigram}$^\dagger$ & $98.03{\pm}.05$ & $96.41{\pm}.04$ & $97.17{\pm}.03$ & $95.25{\pm}.05$ \\
\sys{Mean-Field}$^\dagger$ & $98.06{\pm}.05$ & $96.43{\pm}.04$ & $97.18{\pm}.03$ & $95.29{\pm}.05$ \\
\sys{Diffusion-Uni}$^\dagger$ & $98.25{\pm}.08$ & $96.72{\pm}.05$ & $97.39{\pm}.03$ & $95.68{\pm}.08$ \\
\sys{Diffusion-MF}$^\dagger$ & $\mathbf{98.27{\pm}.04}$ & $\mathbf{96.77{\pm}.06}$ & $\mathbf{97.40{\pm}.05}$ & $\mathbf{95.71{\pm}.07}$ \\
\midrule
\multicolumn{5}{l}{\emph{Learned-transition Viterbi}} \\
\sys{LCRF}$^\dagger$ & $98.02{\pm}.04$ & $96.42{\pm}.04$ & $97.18{\pm}.04$ & $95.26{\pm}.05$ \\
\sys{Diffusion-LCRF}$^\dagger$ & $98.10{\pm}.14$ & $96.58{\pm}.13$ & $97.23{\pm}.10$ & $95.38{\pm}.22$ \\
\bottomrule
\end{tabular}
\end{adjustbox}
\caption{Chinese GSDSimp test results with
\texttt{hfl/chinese-roberta-wwm-ext}.  Values are mean $\pm$ standard
deviation over four seeds; $^\dagger$ marks the development-selected decoder.}
\label{tab:segpos-pretrained-full}
\end{table}

\FloatBarrier
\section{Additional Analyses}
\label{sec:additional-analyses}

\subsection{Model Scaling}
\label{sec:ablation-studies}

Table~\ref{tab:scaling_testing} varies model capacity on English POS.  The
conventional systems scale their scratch encoder, whereas diffusion systems
keep that encoder fixed and allocate additional parameters to the denoising
Transformer (\S\ref{sec:hyperparameters}).  The diffusion models continue to
improve with capacity after the conventional systems saturate.

\begin{table}
\centering
\small
\setlength\tabcolsep{4.5pt}
\begin{tabular}{@{}lccccc@{}}
\toprule
Model & \textsc{XS} & \textsc{S} & \textsc{M} & \textsc{L} & \textsc{XL} \\
\midrule
\multicolumn{6}{l}{\emph{Scratch encoder}} \\
\sys{Unigram}       & 91.17 & 91.36 & 91.16 & 90.87 & -- \\
\sys{Mean-Field}    & 93.17 & 92.96 & 92.79 & 92.63 & -- \\
\sys{LCRF}          & 93.84 & 93.58 & 93.55 & 93.47 & -- \\
\sys{Diffusion-Uni} & --    & 94.04 & 94.33 & 94.47 & 94.83 \\
\sys{Diffusion-MF}  & --    & \textbf{94.45} & \textbf{94.67} & \textbf{94.72} & \textbf{94.97} \\
\midrule
\multicolumn{6}{l}{\emph{RoBERTa-large}} \\
\sys{Unigram}       & 98.33 & \multicolumn{4}{c}{--} \\
\sys{Mean-Field}    & 98.20 & \multicolumn{4}{c}{--} \\
\sys{LCRF}          & 98.22 & \multicolumn{4}{c}{--} \\
\sys{Diffusion-Uni} & --    & 98.37 & 98.40 & 98.42 & \textbf{98.49} \\
\sys{Diffusion-LCRF} & --   & --    & --    & 98.44 & 98.40 \\
\sys{Diffusion-MF}  & --    & \textbf{98.44} & \textbf{98.48} & \textbf{98.47} & \textbf{98.49} \\
\bottomrule
\end{tabular}
\caption{English EWT development POS accuracy on UD~2.15 as model capacity
varies.  Values are means over four seeds; -- denotes an unevaluated setting.}
\label{tab:scaling_testing}
\end{table}


\FloatBarrier
\subsection{Diffusion Hyperparameters}
\label{sec:diffusion_tuning}

\begin{table}
\centering
\small
\setlength\tabcolsep{6pt}
\begin{tabular}{@{}lcccc@{}}
\toprule
Layers & 1 & 2 & 3 & 4 \\
\midrule
\sys{Diffusion-Uni} & 93.76 & 94.45 & 94.62 & 94.72 \\
\sys{Diffusion-MF}  & 94.33 & 94.78 & 94.88 & 94.88 \\
\bottomrule
\end{tabular}

\begin{tabular}{@{}lcccc@{}}
\toprule
Layers & 6 & 8 & 10 & 12 \\
\midrule
\sys{Diffusion-Uni} & 94.80 & 94.84 & 94.86 & \textbf{94.90} \\
\sys{Diffusion-MF}  & \textbf{94.97} & 94.94 & 94.93 & 94.93 \\
\bottomrule
\end{tabular}
\caption{English EWT development POS accuracy as denoiser depth varies, using
the scratch encoder and \textsc{XL} diffusion configuration.  Values are means
over four seeds.}
\label{tab:layers_testing}
\end{table}


This controlled English POS sweep varies only denoiser depth.  Within the
tested range, \sys{Diffusion-Uni} continues to improve through 12 layers,
whereas \sys{Diffusion-MF} peaks at six and then changes little.  These results
motivate treating depth as a development-selected hyperparameter rather than
using one depth for every task.

\paragraph{Mean-Field iterations.}
In a controlled English EWT POS development sweep, additional Mean-Field
updates gave diminishing gains; development selection chose 10 iterations for
\sys{Mean-Field} and 15 within each \sys{Diffusion-MF} denoising call.
Table~\ref{tab:mf_iter_timing} shows the corresponding throughput trade-off.
This sweep isolates iteration count in one scratch POS configuration; the
matched three-task benchmark is reported in Table~\ref{tab:efficiency}.

\begin{table}
\centering
\small
\setlength\tabcolsep{2.2pt}
\begin{tabular}{@{}lrrrrrrr@{}}
\toprule
MF iterations & 1 & 2 & 3 & 4 & 7 & 10 & 15 \\
\midrule
\multicolumn{8}{l}{\emph{Training}} \\
\sys{Mean-Field}   & 100.0 & 99.4 & 97.0 & 97.0 & 93.7 & 89.1 & 85.9 \\
\sys{Diffusion-MF} & 100.0 & 100.0 & 98.2 & 98.2 & 92.8 & 90.1 & 83.5 \\
\midrule
\multicolumn{8}{l}{\emph{Inference}} \\
\sys{Mean-Field}   & 100.0 & 99.7 & 89.1 & 83.6 & 92.8 & 87.2 & 83.2 \\
\sys{Diffusion-MF} & 100.0 & 99.0 & 96.9 & 94.3 & 92.2 & 86.0 & 81.9 \\
\bottomrule
\end{tabular}
\caption{Relative throughput (\%) in the English EWT POS iteration sweep,
normalised to one Mean-Field iteration within each row (higher is better).}
\label{tab:mf_iter_timing}
\end{table}


\FloatBarrier
\subsection{Diffusion Objective and Inference Budget}
\label{sec:diffusion-component-ablation}

We conduct controlled diffusion-specific ablations of scratch
\sys{Diffusion-MF} on English EWT POS, CoNLL-2003 NER, and Chinese Seg+POS,
using seeds 1--3.  For the objective ablation, we retrain each selected
configuration with the full hybrid objective
$\mathcal L_{\mathrm{den}}+\mathcal L_{\mathrm{post}}$, with direct clean-label
denoising removed, or with posterior matching removed.  For the inference
ablation, we select one full-objective checkpoint per seed on development data
and hold it fixed while evaluating 5, 10, or 20 approximately log-spaced
denoiser calls.  POS uses argmax, while NER and Chinese use Mask-Viterbi.
Ten calls exactly reproduce the base-2 halving trajectory used in the main
experiments.  Table~\ref{tab:diffusion-component-ablation} reports test means
and sample standard deviations; within each block, only the component under
study changes.

\paragraph{Contribution of the objective terms.}
Measured as paired full-minus-ablated test differences, removing direct
clean-label denoising changes POS accuracy, NER F1, and Chinese joint F1 by
$0.75{\pm}.10$, $1.23{\pm}.55$, and $0.62{\pm}.33$ points, respectively;
removing posterior matching changes them by
$0.05{\pm}.09$, $0.36{\pm}.69$, and $0.14{\pm}.30$.  Positive values favour the full
objective.  It beats both ablations on average on all three tasks,
and direct denoising has the larger contribution on all three tasks.
The matched-seed variation therefore supports the hybrid objective as a robust
default without implying that the two terms contribute equally on every task.

\paragraph{Number of denoiser calls.}
Relative to ten calls, five calls change POS accuracy, NER F1, and Chinese
joint F1 by $0.03{\pm}.05$, $0.66{\pm}.47$, and
$-0.17{\pm}.03$ points; twenty calls change them by
$-0.01{\pm}.03$, $0.54{\pm}.54$, and
$-0.09{\pm}.08$.  No call budget has the highest mean on every task, and increasing the number of reverse updates gives no monotonic improvement.
Thus five calls provide a lower-cost alternative whose suitability is
task-dependent, while the base-2 ten-call trajectory remains a reasonable
task-independent default.

\FloatBarrier
\subsection{Efficiency Benchmark}
\label{sec:timing-experiments}

We benchmark the three \textsc{L} diffusion denoisers on all tasks using the
same scratch Transformer--char-LSTM encoder, 5,000-token batch budget, and one
NVIDIA A100-SXM4-80GB GPU.  Training throughput covers a complete parameter
update; inference covers the encoder, reverse process, and common argmax, but
excludes loading, metrics, and structured final decoding.  \sys{Diffusion-MF}
uses 15 Mean-Field updates per call and \sys{Diffusion-LCRF} exact first-order
marginals.  After one warm-up, Table~\ref{tab:efficiency} reports the median of
five runs on the training split for training and the development split for
inference.

Table~\ref{tab:speed-accuracy-tradeoff} separately varies the reverse-schedule
base for lightweight English POS denoisers.  A larger base makes fewer calls
and raises throughput while development accuracy remains nearly unchanged.
Together with Table~\ref{tab:diffusion-component-ablation}, this shows that
fewer reverse updates can reduce cost without a consistent accuracy loss; the
sweep is diagnostic rather than a universal operating point.

\begin{table}
\centering
\small
\setlength{\tabcolsep}{2.1pt}
\begin{tabular}{@{}lcrrr@{}}
\toprule
System & $b$ & Dev acc. & Train & Inference \\
       &     &          & \multicolumn{2}{c}{sentences/s} \\
\midrule
\sys{Unigram}   & -- & 98.29 & 181 & 672 \\
\sys{Mean-Field}& -- & 98.13 & 165 & 648 \\
\sys{LCRF}      & -- & 98.28 & 133 & 324 \\
\midrule
\multirow{4}{*}{\sys{Diffusion-Uni} D1}
 & 1.5 & 98.41 & 164 & 267 \\
 & 2   & 98.42 & 162 & 328 \\
 & 3   & 98.42 & 162 & 383 \\
 & 4   & 98.41 & 164 & 422 \\
\midrule
\multirow{4}{*}{\sys{Diffusion-MF} D1/NMF1}
 & 1.5 & 98.47 & 163 & 236 \\
 & 2   & 98.46 & 158 & 289 \\
 & 3   & 98.42 & 164 & 352 \\
 & 4   & 98.42 & 162 & 392 \\
\midrule
\multirow{4}{*}{\sys{Diffusion-LCRF} D1}
 & 1.5 & 98.42 & 68 & 28 \\
 & 2   & 98.43 & 71 & 44 \\
 & 3   & 98.36 & 71 & 57 \\
 & 4   & 98.36 & 67 & 67 \\
\bottomrule
\end{tabular}
\caption{Controlled English EWT POS speed--accuracy sweep.
$b$ is the logarithmic reverse-schedule base; larger values use fewer
denoiser calls.  D1 uses one denoising Transformer layer and NMF1 one
Mean-Field iteration.  Throughput is comparable within this controlled sweep,
but not directly with the rerun in Table~\ref{tab:efficiency}.}
\label{tab:speed-accuracy-tradeoff}
\end{table}


\FloatBarrier
\section{Detailed Related Work}
\label{sec:related}

\paragraph{Diffusion for discrete and structured outputs.}
Direct discrete diffusion defines a finite-state corruption process in the
categorical output space \citep{hoogeboom-2021-argmax-flows,austin-2021-struc-denois}.
Continuous approaches instead embed discrete symbols in a real-valued space,
apply Gaussian diffusion, and discretise the result at decoding time
\citep{ho-2020-denois-diffus,Li2022DiffusionLMIC,Peebles2022DiT}.  Related
variants use iterative masking \citep{Chang2022MaskGITMG}, bit encodings
\citep{Chen2022AnalogBG}, or vector-quantised codebooks \citep{Gu2021VectorQD};
continuous-time categorical formulations connect discrete diffusion to jump
processes and score matching \citep{Sun2023CategoricalDiffusion}.

For structured NLP outputs, \sys{DiffusionNER} denoises entity-boundary tuples
\citep{Shen2023}, while \sys{DiffusionSL} maps token labels to binary codes and
diffuses their continuous relaxation \citep{Huang2023}.  Diffusion has also
been used for non-autoregressive generation with structural controls
\citep{Gong2022DiffuSeqST}.  Other adjacent work places a continuous LCRF around
latent diffusion \citep{Ranasinghe2024LatentCRF} or augments a BiLSTM--LCRF with
diffusion for NER \citep{Qiu2025DiffusionCRF}.  In contrast, our corruption
process is categorical and the denoiser itself performs structured inference
at every reverse step.  To our knowledge, this is the first discrete diffusion
sequence labeller with an explicit structured denoiser.  Exact LCRF denoising
defines normalised sequence scores, while Mean-Field supplies parallel
approximate marginals.

\paragraph{Structured prediction with neural LCRFs.}
LCRFs remain a standard sequence-labelling model
\citep{lafferty-2001-condit-random-field}, but first-order factors directly
couple only adjacent labels.  Neural unrolling makes approximate inference
differentiable \citep{zheng-2015-condit-random}, and parallel methods improve
training or decoding efficiency
\citep{wang-etal-2020-ain,corro-etal-2025-bregman}.  This line connects to
amortized and learned inference
\citep{Stoyanov2011ApproxInferencePolicies,Domke2012,Hershey2014DeepUnfolding}
and to classical variational analyses
\citep{Wainwright2008,Yedidia2005,Murphy1999Loopy}.  Whereas these models
perform one structured prediction pass, our diffusion trajectory repeatedly
conditions each pass on an evolving whole-sequence hypothesis.  This is also
related to the structured denoising view of \citet{jayasumana-2024-markov}, but
we place the LCRF inside a diffusion denoiser for sequence labelling rather than
inside a generative Transformer.


\FloatBarrier


\end{document}